\documentclass[10pt,twocolumn,letterpaper]{article}

\usepackage{cvpr}              
\usepackage{tabularx}  
\usepackage{booktabs} 
\usepackage{amssymb}   
\usepackage{amsmath}
\usepackage{multirow}
\usepackage[table]{xcolor} 
\usepackage{makecell}
\usepackage{array}
\usepackage{algorithm}
\usepackage{algpseudocode}
\usepackage{xcolor}
\definecolor{green}{RGB}{0,255,0}
\definecolor{red}{RGB}{255,0,0}
\definecolor{blue}{RGB}{0,0,255}
\usepackage{pifont}
\usepackage[table]{xcolor}

\usepackage{xcolor} 
\newcommand{\method}{ELF-VLA}

\definecolor{myblue}{HTML}{4486C1}

\definecolor{cvprblue}{rgb}{0.21,0.49,0.74}
\usepackage[pagebackref,breaklinks,colorlinks,allcolors=cvprblue]{hyperref}

\def\paperID{40802} 
\def\confName{CVPR}
\def\confYear{2026}

\title{Unleashing VLA Potentials in Autonomous Driving via \\  Explicit Learning from Failures} 

\author{
    Yuechen Luo\textsuperscript{1*},
    Qimao Chen\textsuperscript{1*},
    Fang Li\textsuperscript{2*},  
    Shaoqing Xu\textsuperscript{2,\textdaggerdbl}, 
    Jiaxin Liu\textsuperscript{1}, 
    Ziying Song\textsuperscript{3}, \\
    Zhi-xin Yang\textsuperscript{2,\ding{41}},
    Fuxi Wen\textsuperscript{1,\ding{41}}\\
\textsuperscript{1} Tsinghua University, \textsuperscript{2} University of Macau, \textsuperscript{3} Beijing Jiaotong University \\
{\tt\small luo-yc24@mails.tsinghua.edu.cn}
}

\begin{document}
\maketitle

\let\thefootnote\relax\footnotetext{*Equal contribution. \textdaggerdbl Project Leader. \ding{41} Corresponding author. }

\begin{abstract}
Vision-Language-Action (VLA) models for autonomous driving often hit a performance plateau during Reinforcement Learning (RL) optimization. This stagnation arises from exploration capabilities constrained by previous Supervised Fine-Tuning (SFT), leading to ``persistent failures" in long-tail scenarios. In these critical situations, all explored actions yield a zero-value driving score. This information-sparse reward signals a failure, yet fails to identify its root cause—whether it is due to incorrect planning, flawed reasoning, or poor trajectory execution. 
To address this limitation, we propose \textbf{VLA} with \textbf{E}xplicit \textbf{L}earning from \textbf{F}ailures (\textbf{\method}), a framework that augments RL with structured diagnostic feedback. Instead of relying on a vague scalar reward, our method produces detailed, interpretable reports that identify the specific failure mode. 
The VLA policy then leverages this explicit feedback to generate a \textbf{Feedback-Guided Refinement}. By injecting these corrected, high-reward samples back into the RL training batch, our approach provides a targeted gradient, which enables the policy to solve critical scenarios that unguided exploration cannot. Extensive experiments demonstrate that our method unlocks the latent capabilities of VLA models, achieving state-of-the-art (SOTA) performance on the public NAVSIM benchmark for overall PDMS, EPDMS and high-level planning accuracy. 
\end{abstract}    
\section{Introduction}
\label{sec:intro}

The development of autonomous driving systems is undergoing a paradigm shift from traditional modular architectures to end-to-end frameworks~\cite{e2esurvey, hu2023planning}. Vision-Language-Action (VLA) models are at the forefront of this transition~\cite{vlasurvey}. These models map raw camera sensor inputs to coherent vehicle motion commands by applying supervised fine-tuning (SFT) and reinforcement learning (RL) to large Vision-Language Models (VLMs). This integrated design eliminates manually engineered interfaces and supports large-scale, data-driven policy learning. Notably, VLA models can generate intermediate reasoning trajectories via a ``think" module, mimicking human problem-solving strategies, offering a promising direction toward achieving explainable and trustworthy autonomous driving~\cite{zhou2025autovla,adathinkdrive, driver1}.

\begin{figure}[t!]
    \centering
    \includegraphics[width=\linewidth]{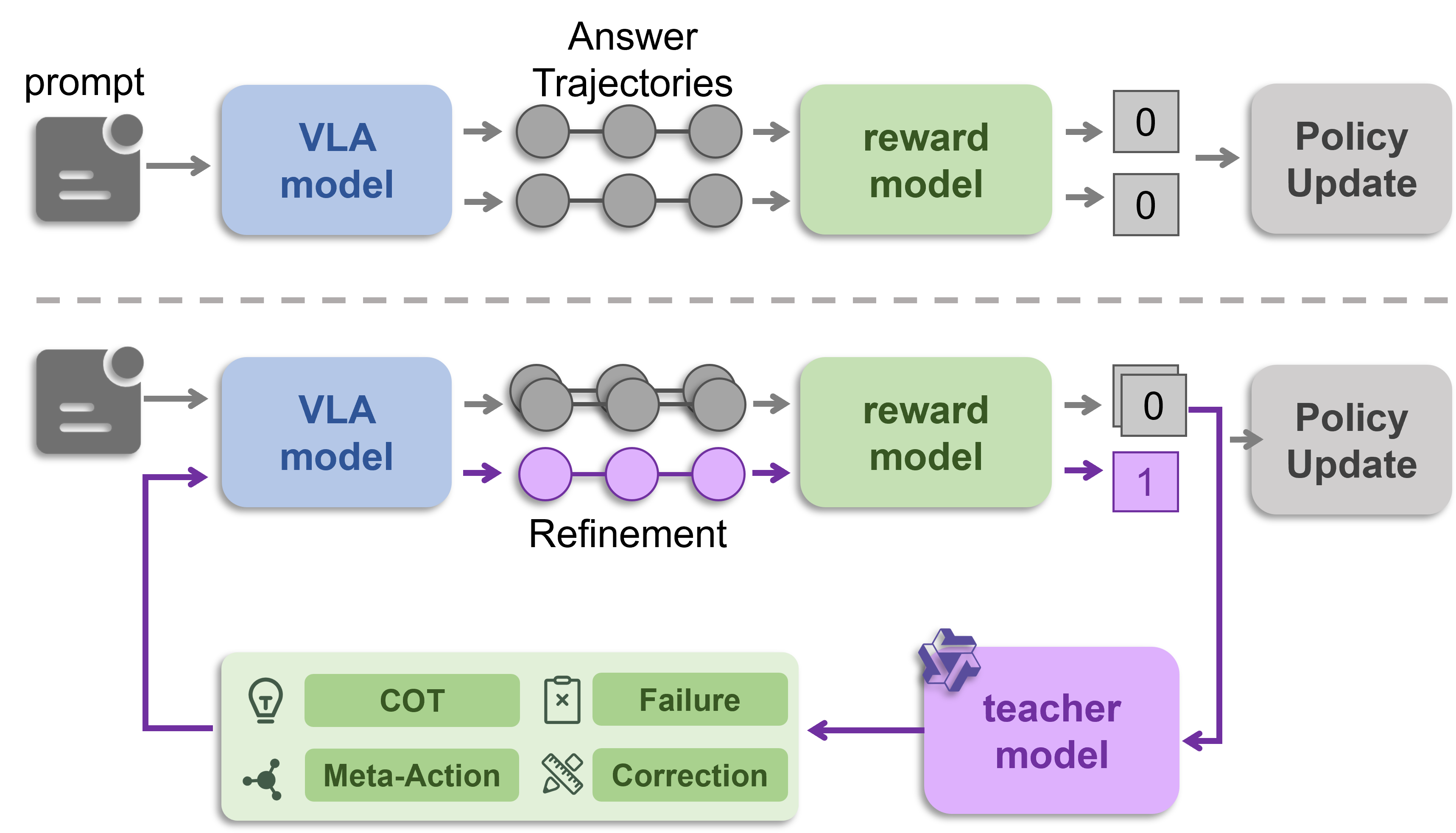}
    \caption{\textbf{The comparison between RL fine-tuning of general VLA and \method.} Top: VLA training with RL algorithm suffers from a performance plateau: in certain scenarios, the policy model's rollouts consistently yield low-scoring answers, trapping the agent and preventing it from discovering a better policy. Bottom: \method\  addresses this by using a teacher model to provide structured feedback, which is then used to re-rollout a refinement, forcing the policy to break through this performance plateau.}
    \vspace{-5mm}
    \label{fig:intro}
\end{figure}

Despite this progress, RL fine-tuning continues to exhibit a performance plateau: 
we observe that after SFT, the model's policy exploration capability is severely constrained by the SFT dataset's limitations, where common scenarios are highly prevalent and the safety-critical scenarios that rigorously test the autonomous system's capabilities are rare~\cite{curseofrarity, safetycriticalsurvey}. Consequently, under safety-critical and challenging scenarios (such as complex unprotected left turns or emergency evasions), all exploratory rollouts consistently fail, yielding a zero driving score, as shown in the top row of Fig.~\ref{fig:intro}.
Existing VLA-RL approaches simplify performance evaluation during training to a single scalar reward (e.g., PDMS~\cite{dauner2024navsim}). When the model fails, this information-sparse reward is insufficient to pinpoint the root cause of the error, making it unclear whether the failure stems from the cumulative errors of high-level planning in the ``think" module, flawed cognitive reasoning about critical targets, or dynamic deficiencies in the low-level trajectory.

To address these limitations and enable continuous learning, this paper proposes a novel VLA 
training framework for autonomous driving that bridges failure diagnosis and policy correction. As shown in the bottom row of Fig.~\ref{fig:intro}, the main idea is to provide feedback with structured failure analysis to help VLA with its ``Think-then-Act" architecture, rather than relying on simple scalar rewards. This approach features two core innovations:

\begin{itemize}
    \item \textbf{VLA Ability-aligned Feedback:} We introduce a feedback mechanism using a teacher model, which is triggered when the VLA encounters persistent failures. This model generates a structured diagnostic report aligned with VLA's ability that pinpoints specific errors within the VLA's planning, reasoning, or execution levels.
    \item \textbf{Feedback-Guided Refinement and Re-injection:} The VLA policy model (student) leverages this diagnostic report to generate a corrected trajectory. This high-reward corrected sample is then re-injected into the GRPO training batch. This process provides a goal-directed gradient signal that was previously non-existent in the rollout batch.
\end{itemize}

Through extensive evaluations on the Navsim benchmark, our method demonstrates significant performance improvements over existing VLA baselines. 
Our approach achieves SOTA performance on both the overall driving metric (PDMS) and high-level planning accuracy. By bridging explainable feedback with policy correction, our work provides a practical path for VLA models to overcome performance plateaus in autonomous driving.

\section{Related Work}
\label{sec:relatedwork}

\noindent \textbf{VLA models for autonomous driving.}
The integration of visual and textual data for unified perception, planning, and decision-making has led to a surge of interest in Vision-Language Models (VLMs) for autonomous driving in recent years. Two primary paradigms currently dominate the field. The first is dedicated to scene understanding and high-level reasoning ~\cite{jiang2024senna,jiang2025alphadrive,marcu2024lingoqa,tian2024drivevlm,mtrdrive}. An example of this approach, Senna ~\cite{jiang2024senna}, processes sensory inputs to generate meta-actions for downstream planners, yet substantial improvements in actual driving performance have not been fully realized. The alternative paradigm concentrates on the direct prediction of driving trajectories from raw inputs ~\cite{hwang2024emma,xing2025openemma,qiao2025lightemma,zhao2025sce2drivex,liu2025reasonplan,wang2025omnidrive,fu2025orion, li2025recogdrive}. A notable development, aimed at enhancing model interpretability and accuracy, is the increasing use of intermediate reasoning (e.g., Chain-of-Thought, CoT) to reveal internal cognitive processes. Evidence from EMMA ~\cite{hwang2024emma}, ReasonPlan ~\cite{liu2025reasonplan}, and Sce2DriveX ~\cite{zhao2025sce2drivex} confirms that domain-specific reasoning markedly improves the precision of trajectory forecasting.

\noindent \textbf{RL fine-tuning for VLA models.}
Currently, VLA models in autonomous driving are typically trained using a two-stage paradigm: an initial supervised fine-tuning (SFT) phase on a driving dataset, followed by a reinforcement learning (RL) training phase. In this framework, the efficacy of the RL stage is heavily dependent on the performance of the preceding SFT. 
Current approaches~\cite{adathinkdrive, driver1, zhou2025autovla} employ the Group Relative Policy Optimization (GRPO)~\cite{shao2024deepseekmath} algorithm for RL training, where rewards are measured by VLA's driving score (e.g., PDMS). 
This leads to a significant training deficiency: the VLA model, following SFT phase, struggles to handle the rare, long-tail scenarios present in the training dataset. Consequently, when the model enters RL stage, its driving score in these specific scenarios remains extremely low, regardless of the number of rollouts. This causes the model's overall learning to stagnate, resulting in a performance plateau.
In LLM literature, approaches have successfully used non-numerical feedback, such as textual critiques, to provide detailed guidance~\cite{zhang2025critique,chen2024learning}.  Others~\cite{luffy,ma2025learning} have used mix-policy methods to internalize knowledge from high-quality data, enhancing exploration and policy quality.
Inspired by these approaches, our work introduces a feedback mechanism to address this limitation in the autonomous driving domain, which employs a teacher model to analyze and rectify the VLA model's erroneous driving behaviors by structured feedback, correcting them into proper actions. 

\section{Methods}
\label{sec:methods}

\begin{figure*}[htb]
    \centering
    \includegraphics[width=\linewidth]{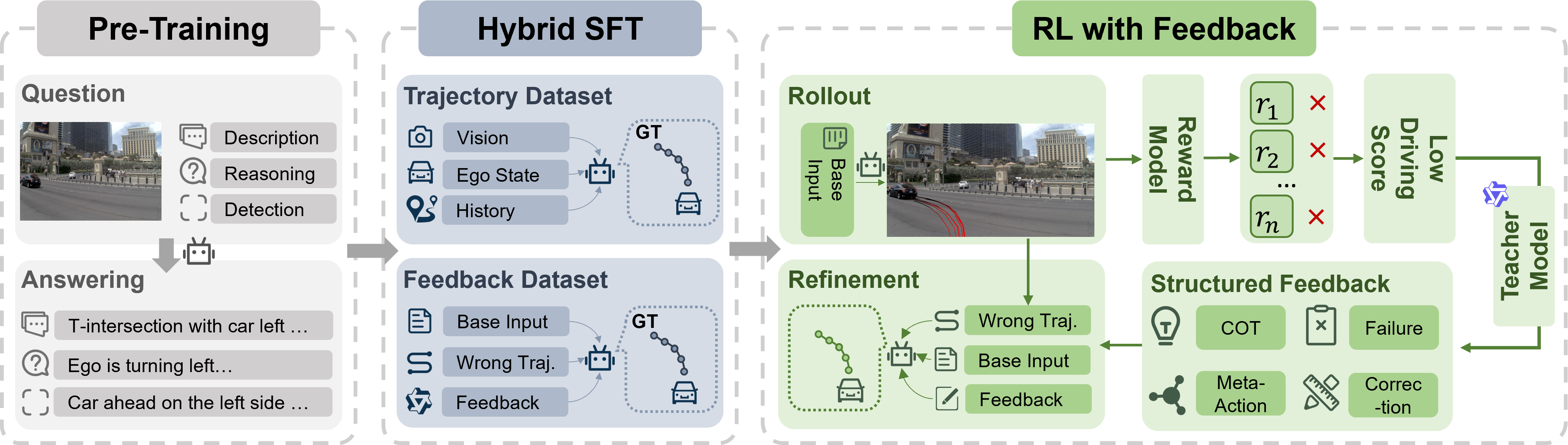}
    \caption{\textbf{Overview of \method.} First, the model is pre-trained on an autonomous driving Q\&A dataset to provide it with foundational driving knowledge. Subsequently, it undergoes SFT on a mixed dataset of ``Base Inputs" and ``Feedback Inputs", enabling it to learn trajectory prediction and feedback-based refinement simultaneously. Finally, in the RL phase, a teacher model is used to generate feedback, thereby reducing the proportion of zero-reward rollouts.}
    \label{fig:overview}
    \vspace{-5mm}
\end{figure*}
In this section, we present our proposed method (Fig.~\ref{fig:overview}), which contains two main components: (1) a two-stage Supervised fine-Tuning (SFT) process, and (2) a Reinforcement Learning (RL) framework enhanced with failure feedback.

\subsection{VLA Inputs Formulation} \label{formulation}
In our method, the VLA model is as the generator and refiner simultaneously, designed to accept two distinct types of inputs: the original, feedback-free \textbf{base inputs}, and the \textbf{feedback inputs}, which incorporate corrective guidance.

\noindent \textbf{Base inputs.}~~
The base input query \( q^{base} \) includes a front-view image denoted as \( I_{\text{cam}} \), high-level navigation commands \( q_{\text{com}} \) (e.g., \textit{Move Forward, Turn Left, Turn Right}), ego state information \( q_{\text{ego}} \) (e.g., \textit{velocity and acceleration}), and the historical trajectory of the last three frames \( q_{\text{his}} = \{h_{t-3}, h_{t-2}, h_{t-1}\} \) at a frequency of 2Hz.

\noindent \textbf{Feedback inputs.}\label{fb}
Based on the base inputs, the VLA model outputs an original response $o$ consisting of trajectory with CoT (details in Appendix~\ref{cot}), which is then classified based on a threshold $s$: responses with a PDMS exceeding $s$ are deemed ``correct" ($o_c$), while those with a score below $s$ are deemed ``wrong" ($o_w$).
\begin{equation}
     q_{fb} = \begin{cases}
     <q_{base}, o_c, f^{rule}> & \text{if}\ o_c,
     \\<q_{base}, o_w, f^{teacher}> & \text{if}\ o_w.
\end{cases} 
\end{equation}
In the case of a correct response, the corresponding feedback inputs are composed of three components: the original base inputs $q_{base}$, the correct response $o_c$ itself, and a rule-based positive feedback $f^{rule}$.
For wrong responses, we employ an external intervention through a VLM teacher model. This teacher model takes the base inputs $q^{base}$, the erroneous trajectory $o_w$, and the ground-truth trajectory $o_{gt}$ to generate a structured feedback $f^{teacher}$. This feedback includes (1) Meta Action Analysis, (2) Think Process Analysis, (3) Safety Failure Analysis, (4) Efficiency Failure Analysis and (5) Actionable Correction (with lateral and longitudinal components). The final feedback inputs for the VLA are then constructed by combining the base inputs, the original wrong response $o_i$, and the generated structured feedback. Detailed examples of base inputs and feedback inputs are in the Appendix~\ref{appendix:prompt}.

\subsection{Two-Stage SFT for Cognition and Refinement}\label{sft}

We employ a two-stage supervised fine-tuning procedure to develop a model that combines both driving knowledge and trajectory planning capabilities. The first stage is designed to infuse the model with general driving knowledge. The second stage is dedicated to equipping the model with the capacity for trajectory prediction, along with the ability to implement refinements based on received feedback.

As shown in Fig.~\ref{fig:overview}, in the first stage, the model is pretrained on a large dataset of driving-related Q\&A pairs to enhance its understanding of driving domain cognition. This dataset is assembled from a diverse collection of open-source driving QA datasets, including DriveLM~\cite{sima2024drivelm}, LingoQA~\cite{marcu2024lingoqa}, ImpromptuVLA~\cite{chi2025impromptu}, and other open-source driving datasets~\cite{qian2024nuscenes, ding2024holistic,wang2025omnidrive}. In addition, we constructed a multi-turn Q\&A reasoning dataset for NAVSIM following the CoT paradigm. This phase addresses tasks like road boundary estimation (drivable area), critical object identification (object localization), ego action prediction, and related traffic semantics. Further details on dataset composition are provided in the Appendix~\ref{appendix:pretrain_data}.

Subsequently, the second stage introduces the trajectory prediction and refinement task. For each query $q^{base}$ and $q^{fb}$ (defined in Sec.~\ref{grpo}), the model's output is supervised by the ground-truth trajectory $o_{gt}$, aiming to maximize the conditional likelihood:
\begin{equation}
\mathcal{L}_{\text{SFT}} = 
\mathbb{E}_{(q, o) \sim \mathcal{D}} \big[ -\log \pi_\theta(o \mid q) \big].
\end{equation}

$\mathcal{D}$ denotes the dataset including $\{(q_{base}, o_{gt}), (q_{fb}, o_{gt})\}$ and $\pi_{\theta}$ denotes the VLA model. This mixed-dataset training approach equips the model with the dual capabilities of trajectory prediction and feedback-based trajectory refinement, thereby enabling the model to leverage failure feedback during the reinforcement learning phase. 

\subsection{RL with Failure Feedback}
\label{grpo}
\begin{figure*}[t!]
    \centering
    \includegraphics[width=0.95\linewidth]{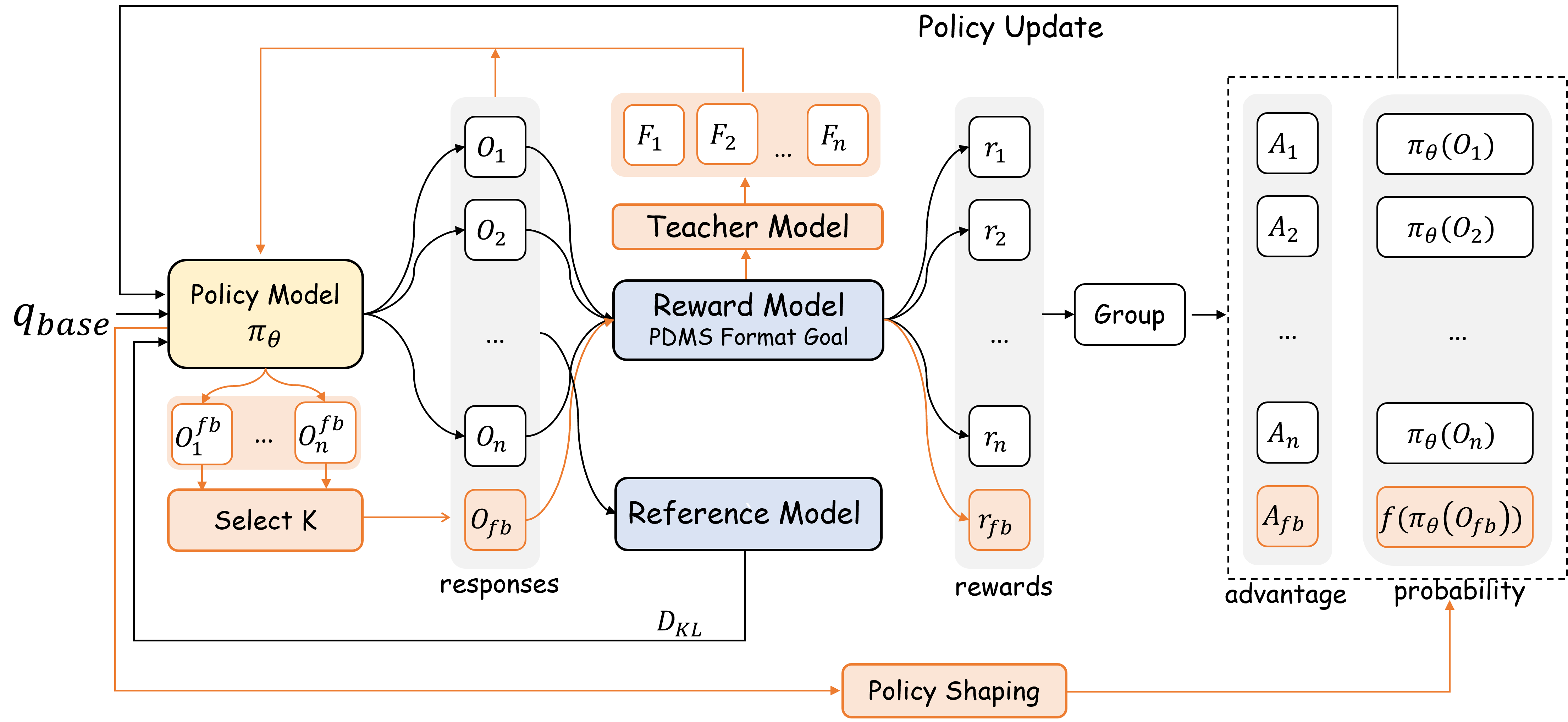}
    \caption{\textbf{Overview of GRPO with feedback}. The policy model generates initial responses. Based on the rewards, teacher model (Qwen3-VL-32B) provides feedback, guiding the policy to sample improved refinement responses. A high-quality refinement response is selected and combined with the initial response set for joint optimization. Policy Shaping is applied to the final probability.}
    \label{fig:grpo_feedback}
\end{figure*}

Our failure-feedback mechanism is applied during the rollout phase of the GRPO algorithm, inspired by \cite{zhang2025critique}.
Conventional VLA models for autonomous driving typically hit a performance bottleneck during RL training. This occurs because they cannot manage complex, long-tail scenarios; consequently, the trajectories sampled in these situations receive extremely low driving scores, resulting in a sparse reward problem.
Our method addresses this by introducing a feedback mechanism that successfully boosts the model's driving scores in these critical scenarios, enabling the agent to break through the performance plateau.

\noindent \textbf{Efficient Difficult-Sample Curation.}
Before introducing the GRPO with a feedback mechanism, we first perform a cost-effective data curation to maximize training efficiency. Naive RL training often wastes resources on overly simple (already mastered) scenarios that provide weak learning signals. Our curation aims to filter out these samples and focus the agent on high-value, informative scenarios, which include both difficult samples (where the model consistently fails) and ambiguous samples (where the model is most uncertain). To achieve this, we utilize the SFT model to sample $N$ rollouts for each query, estimating its mean reward and reward variance. We then discard samples characterized by high mean reward and low variance, as these indicate consistent success. This strategy effectively concentrates training on the difficult (low mean, low variance) and ambiguous (high variance) scenarios. Through this method, we filter the initial 85k training entries down to a core dataset of 24k high-value scenarios.

\noindent \textbf{Reward modeling.} To incentivize the VLA model to learn effective driving behaviors and to ensure the stability of its output format, we designed a reward function with three components: the PDMS Reward, the Format Reward, and the Goal Reward.
The \textbf{PDMS Reward} $r_{traj}$ is a comprehensive trajectory evaluation metric based on the Predictive Driver Model Score~\cite{dauner2024navsim}. It is represented as a continuous value ranging from 0 to 1. The specific formula used to calculate this score is provided in Appendix~\ref{navsim_metric}.
The \textbf{Format Reward} $r_{fmt}$ is a binary (1 or 0) reward designed to enforce adherence to the required output format strictly.
Finally, the \textbf{Goal Reward} $r_{goal}$ incentivizes endpoint accuracy by assigning a tiered reward based on the L1 distance to the GT endpoint. Detailed calculations for each reward are provided in the Appendix~\ref{appendix:reward}.
The overall reward in the reinforcement learning process is computed by integrating four designed reward components, which are presented as follows:
\vspace{-0.5em}
\begin{equation}
    r = r_{traj} + r_{fmt} + r_{goal}.
    \vspace{-0.5em}
\end{equation}

\noindent \textbf{GRPO with feedback.}
Our method employs a feedback mechanism, illustrated in Fig.~\ref{fig:grpo_feedback}, to refine trajectories and increase rewards, thereby enabling the VLA model to surpass its performance plateau.
More specifically, the process begins by sampling a batch of trajectory responses $\{o_i\}_{i=1}^{n}$ using the base inputs $q^{base}$. The rewards for this batch $\{r_i\}_{i=1}^{n}$ are then computed, which include $\{r_{traj,i}\}_{i=1}^{n}$, $\{r_{fmt,i}\}_{i=1}^{n}$, and $\{r_{goal,i}\}_{i=1}^{n}$. Based on the predefined threshold $s$, these responses are then classified into two groups: correct responses $\{o_c\}$ and wrong responses $\{o_w\}$. Finally, following Sec.~\ref{fb}, both the correct and wrong responses are processed accordingly and then assembled to create the final feedback inputs $\{q_{i}^{fb}\}_{i=1}^{n}$.

Subsequently, the VLA model generates a new batch of responses $\{o_{i}^{fb}\}_{i=1}^{n}$ from the feedback inputs ($\{q_{i}^{fb}\}_{i=1}^{n}$) and calculates their rewards $\{r_i^{fb}\}_{i=1}^{n}$. We randomly select $k$ responses from the subset of $\{o_i^{fb}\}_{i=1}^{n}$ whose trajectory rewards ($r_{traj}^{fb}$) exceed the original batch's maximum trajectory reward $max(r_{traj})$. If fewer than $k$ such ``better" responses exist, the remaining slots are filled by duplicating the original response that achieved this maximum reward. This results in a final batch of $n+k$ rollout samples $\{(q_i^{final}, o_i^{final})\}_{i=1}^{n+k}$, from which the relative advantage $\{A_i\}_{i=1}^{n+k}$ is then computed. Details of the rollout update algorithm are in Alg.~\ref{alg:rolloutupdate}. The final optimization objective for GRPO is defined as follows:
\begin{equation}
    \mathcal{J}(\theta) = \mathbb{E}_{D_{final} \sim \pi_{\theta_{old}}} \left[ \frac{1}{n} \sum_{i=1}^{n} \mathcal{J}_i + \frac{1}{k} \sum_{j=1}^{k}\mathcal{J}_j^{fb} - \beta \mathbb{D}_{KL} \right],
\end{equation}
\begin{equation}
    \mathcal{J}_i = \min \big( c_i(\theta) A_i, \text{clip}\left(c_i(\theta), 1 - \epsilon, 1 + \epsilon\right) A_i \big),
\end{equation}
\begin{equation}
    \mathcal{J}_j^{fb} = f(c_j^{fb}(\theta))A_j^{fb},
\end{equation}

\noindent where $D_{final}=\{(q_i^{final}, o_i^{final})\}_{i=1}^{n+k}$, $\pi_{ref}$ is the reference policy of the initial SFT model, and $\beta$ is a hyper-parameter. We apply the CLIP only to the original batch rollout samples, not to the responses that have been refined by feedback. To compute the advantages, we first merge both sets of rewards into a unified set $r_{union}$. The mean and standard deviation of this combined set are then used to normalize the rewards and calculate the relative advantages, $A_i$ and $A_j^{fb}$, as follows:
\begin{equation}
    r_{union} = \{{r_j}\}_{j=1}^n \cup \{{r_{j'}^{fb}}\}_{j'=1}^k,
\end{equation}
\begin{equation}
    A_i = \frac{r_i-\text{mean}(r_{union})}{\text{std}(r_{union})}, A_i^{fb} = \frac{r_j^{fb}-\text{mean}(r_{union})}{\text{std}(r_{union})},
\end{equation}

For the token-level probability ratios of the feedback-generated outputs, a challenge arises from a mismatch in conditioning. These samples are generated using the feedback query $q^{fb}$, but our optimization objective is conditioned on the base query $q^{base}$. This discrepancy can cause the refined responses $\{o^{fb}\}$ to have very low probabilities under the optimization policy, leading to high variance, potential gradient explosion, and training instability. Therefore, inspired by LUFFY~\cite{luffy}, we employ Policy Shaping $f(x)=\frac{x}{x+\gamma}$($0<\gamma<1$). This technique assigns higher weights to low-probability tokens within the $\{o^{fb}\}$. Such mechanism encourages the model to learn valuable knowledge from rare but correct trajectories that might otherwise be overlooked. The standard ratio $c_i(\theta)$ and the shaped ratio $f(c_j^{fb}(\theta))$ are defined as:
\begin{equation}
    c_i(\theta) = \frac{\pi_\theta(o_i |q^{base})}{\pi_{{old}}(o_i |q^{base})} , f(c_j^{fb}(\theta)) = \frac{\pi_\theta(o_j^{fd} |q^{base})}{\pi_{{\theta}}(o_j^{fd} |q^{base})+\gamma}.
\end{equation}

\setlength{\textfloatsep}{0.2cm}
\begin{algorithm}[t]
\caption{Rollout Update of GRPO with Feedback}
\label{alg:rolloutupdate}
\begin{algorithmic}[1]
\small
\linespread{1.3}\selectfont

\Require
\Statex $\{q_{i}^{base}\}_{i=1}^{n}$: Base Inputs
\Statex $s$: Score Threshold  
\Statex $k$: Number of Refinements 

\Ensure
\Statex $\{q_i^{final}\}_{i=1}^{n+k}$: Final Query
\Statex $\{o_i^{final}\}_{i=1}^{n+k}$: Final Rollout Responses

\Statex \textit{// \textbf{Step1: Initial Rollout}}
\State $\{o_i\}_{i=1}^{n} = \{\texttt{VLA-MODEL}(q_i^{base})\}_{i=1}^n$

\State $\{r_{traj,i}\}_{i=1}^{n}=\{r_{traj}(o_i)\}_{i=1}^{n}$
\State $\{f_i^{teacher}\}_{i=1}^{n}=\{\texttt{Teacher-Model}(q_{base}, o_i, o_{gt})\}_{i=1}^n$
\State $\{q_i^{fb}\}_{i=1}^{n}=\begin{cases}
     <q_i^{base},o_i,f^{rule}>,&\text{if}\  r_{traj}(o_i)\ge s\\
     <q_i^{base},o_i,f_i^{teacher}>,&\text{if}\  r_{traj}(o_i)< s
\end{cases}$
\Statex \textit{// \textbf{Step2: Rollout with Feedback}}
\State  $\{o_{i}^{fb}\}_{i=1}^{n}=\{\texttt{VLA-Model}(q_i^{fb})\}_{i=1}^{n}$
\State $\{r_{traj,i}^{fb}\}_{i=1}^{n}=\{r_{traj}(o_i^{fb})\}_{i=1}^{n}$
\Statex \textit{// \textbf{Step3: Final Query and Output Construction}}
\State $r_{\max} \leftarrow \max_{i}(\{r_{traj,i}\}_{i=1}^{n})$
\State $\{o_{j}^{fb}\}_{j=1}^{k} = \texttt{Select}(k, \{o_m^{fb} \mid r_{traj,m}^{fb} > r_{\max}\})$
\State $\{q_i^{final}\}_{i=1}^{n+1} = \{q_i^{base}\}_{i=1}^n+\{q_j^{base}\}_{j=1}^k$
\State $\{o_i^{final}\}_{i=1}^{n+1} = \{o_i\}_{i=1}^n+\{o_j^{fb}\}_{j=1}^k$
\State \Return $\{q_i^{final}\}_{i=1}^{n+k}$, $\{o_i^{final}\}_{i=1}^{n+k}$
\end{algorithmic}
\end{algorithm}

\section{Experiment}
\label{sec:experiment}

\begin{table*}[t]
\renewcommand{\arraystretch}{1}
\begin{center}
\caption{Comparison with state-of-the-art methods on the NAVSIMv1 with PDMS.}
\vspace{-0.5em}
\begin{tabularx}{\textwidth}{l|c|c|XXXXX|c}
\toprule
\textbf{Method} & \textbf{Image} & \textbf{Lidar} & \textbf{NC}$\uparrow$ & \textbf{DAC}$\uparrow$ & \textbf{TTC}$\uparrow$ & \textbf{CF}$\uparrow$ & \textbf{EP}$\uparrow$ & \textbf{PDMS}$\uparrow$ \\
\midrule
Constant Velocity & & & 68.0 & 57.8 & 50.0 & 100 & 19.4 & 20.6 \\
Ego Status MLP & & & 93.0 & 77.3 & 83.6 & 100 & 62.8 & 65.6 \\
\midrule
UniAD ~\cite{hu2023planning} & \ding{51} & & 97.8 & 91.9 & 92.9 & 100 & 78.8 & 83.4 \\
TransFuser ~\cite{chitta2022transfuser} & \ding{51} & \ding{51} & 97.7 & 92.8 & 92.8 & 100 & 84.0 & 84.0 \\
DiffusionDrive ~\cite{liao2025diffusiondrive} & \ding{51} & \ding{51} & 98.2 & 96.2 & 94.7 & 100 & 82.2 & 88.1 \\
WoTE ~\cite{li2025end} & \ding{51} & \ding{51} & 98.5 & 96.8 & 94.9 & 99.9 & 81.9 & 88.3 \\
Hydra-NeXt ~\cite{li2025hydra} & \ding{51} & & 98.1 & 97.7 & 94.6 & 100 & 81.8 & 88.6\\
AutoVLA-3B~\cite{zhou2025autovla} & \ding{51} & & 98.4 & 95.6 & 98.0 & 100 & 81.9 & 89.1 \\
DriveVLA-W0-3B~\cite{xing2025goalflow} & \ding{51} &  & 98.7 & \textbf{99.1} & 95.3 & 99.3 & 83.3 & 90.3 \\
GoalFlow~\cite{xing2025goalflow} & \ding{51} & \ding{51} & 98.4 & 98.3 & 94.6 & 100 & 85.0 & 90.3 \\
\midrule
 InternVL3-8B-SFT & \ding{51} & & 98.5 & 95.5 & 95.3 & 100 & 81.2 & 87.4 \\
 InternVL3-8B-RL & \ding{51} & & 98.5 & 96.7 & 95.4 & 100 & 83.2 & 89.0  \\
 \rowcolor{gray!30} \textbf{\method-8B}(Ours) & \ding{51} & & \textbf{98.9} & 98.1 & \textbf{96.0} & \textbf{100} & \textbf{85.3} & \textbf{91.0} \\
\bottomrule
\end{tabularx}
\label{table:main_table}
\vspace{-1.5em}
\end{center}
\end{table*}

\begin{table*}[t!]
\centering
\caption{Comparison with state-of-the-art methods on the NAVSIMv2 with EPDMS. 
}
\vspace{-0.5em}
\label{table:navsimv2}
\resizebox{\textwidth}{!}{
\begin{tabular}{l|ccccc|cccc|c} 
\toprule
\textbf{Method} & \textbf{NC $\uparrow$} & \textbf{DAC $\uparrow$} & \textbf{DDC $\uparrow$} & \textbf{TLC $\uparrow$} & \textbf{EP $\uparrow$} & \textbf{TTC $\uparrow$} & \textbf{LK $\uparrow$} & \textbf{HC $\uparrow$} & \textbf{EC $\uparrow$} & \textbf{EPDMS $\uparrow$} \\
\midrule
HydraMDP++~\cite{li2024hydra} & 97.2 & 97.5 & 99.4 & 99.6 & 83.1 & 96.5 & 94.4 & 98.2 & 70.9 & 81.4 \\
DriveSuprem~\cite{yao2025drivesuprim} & 97.5 & 96.5 & 99.4 & 99.6 & 88.4 & 96.6 & 95.5 & \textbf{98.3} & 77.0 & 83.1 \\
Recogdrive-8B~\cite{li2025recogdrive} & 98.3 & 95.2 & \textbf{99.5} & \textbf{99.8} & 87.1 & 97.5 & 96.6 & \textbf{98.3} & 86.5 & 83.6 \\
DiffusionDrive~\cite{liao2025diffusiondrive} & 98.2 & 95.9 & 99.4 & \textbf{99.8} & 87.5 & 97.3 & 96.8 & \textbf{98.3} & \textbf{87.7} & 84.5 \\
DriveVLA-W0-3B~\cite{li2025drivevla} & 98.5 & \textbf{99.1} & 98.0 & 99.7 & 86.4 & 98.1 & 93.2 & 97.9 & 58.9 & 86.1 \\
\midrule
\rowcolor{gray!30} \textbf{\method-8B}(Ours) & \textbf{98.9} & 98.1 & 99.4 & \textbf{99.8} & \textbf{88.5} & \textbf{98.4} & \textbf{96.9} & \textbf{98.3} & 87.2 & \textbf{87.1} \\
\bottomrule
\end{tabular}
}
\vspace{-3mm}
\end{table*}

\subsection{Implementation details}
\noindent \textbf{Dataset.}
We perform comprehensive experiments and evaluations on NAVSIM~\cite{dauner2024navsim}, a planning-oriented autonomous driving dataset built on the OpenScene. In addition to reasoning data collected from NAVSIM, we also leverage several open-source datasets, as described in Sec.~\ref{sft}.

\noindent \textbf{Metric.}\label{metric} 
We evaluate our method's performance on two distinct aspects of autonomous driving: \textbf{high-level planning} and \textbf{trajectory prediction}.

For high-level planning evaluation, we use High-Level Planning Accuracy, which strictly requires the entire meta-action, comprising both longitudinal speed and lateral path, to match the ground truth exactly. Here, the ground truth is generated by GT trajectory, detailed infos in Appendix~\ref{appendix:high_level}.

For trajectory prediction evaluation on the NAVSIM benchmark, we utilize the Predictive Driver Model Score (PDMS) for \textbf{NAVSIMv1}~\cite{dauner2024navsim} and the Extended Predictive Driver Model Score (EPDMS) for \textbf{NAVSIMv2}~\cite{cao2025pseudo} as the closed-loop planning metrics.

\noindent \textbf{Training Details.}
We use InternVL3-8B ~\cite{zhu2025internvl3} as the base model, trained in three stages. First, we pretrain on a large-scale driving knowledge dataset. Second, we fine-tune the model on a hybrid dataset, consisting of curated Navsim planning dataset (with CoT annotations) and the feedback dataset from Sec.~\ref{sec:methods}. Third, we apply reinforcement learning using 32 NVIDIA H20 GPUs. We use Qwen3-VL-32B ~\cite{bai2025qwen2} as the teacher model. Key RL parameters include 8 rollouts per batch, a threshold $s=0.8$, a policy shaping parameter $\gamma=0.1$, and $k=1$ refinement response. Additional details and hyperparameters are in the Appendix~\ref{appendix:training}.

\subsection{Performance Comparison} \label{exp_compa}
\noindent \textbf{Navsim Benchmark.}
Tab.~\ref{table:main_table} presents the performance comparison between \method~and current leading methods on the NAVSIMv1 benchmark. Under the vision-only setting, \method~achieves a PDMS of 91.0, establishing a new state-of-the-art (SOTA). This result represents a significant improvement of 0.7 PDMS over the previous best vision-only method, DriveVLA. Furthermore, \method~outperforms the SFT-only (InternVL3-8B-SFT) and traditional RL (InternVL3-8B-RL) baselines by 3.6 and 2.0 PDMS, respectively.
On the NAVSIMv2 benchmark (Tab.~\ref{table:navsimv2}), \method~ continues its strong performance by achieving a new SOTA of 87.1 EPDMS. This score surpasses the previous best from DriveVLA-W0 by 1.0 EPDMS.
These findings demonstrate that our method \method~substantially enhances the model's driving capabilities over conventional RL approaches, particularly in addressing challenging driving scenarios. Moreover, the outstanding performance across both benchmarks confirms that \method~ is not merely overfitting to the PDMS metric; rather, it exhibits robust generalization by excelling on the distinct and more comprehensive EPDMS as well. 

\noindent{\textbf{Quantitative Evaluation.}} 
We compare the performance of \method~(Tab.~\ref{table:table2}) against several carefully designed ablation models~(detailed definitions in Appendix~\ref{appedix:fb_mechanisms}):
\begin{itemize}  
  \item \textbf{SFT (Baseline):} The base model trained solely with Supervised Finetuning.   
  \item \textbf{GRPO:} The SFT model was further finetuned using the conventional GRPO algorithm. 
  \item \textbf{GT-GRPO:} The SFT model finetuned on a response set augmented with Ground Truth (GT) trajectories, which are added directly.
  \item \textbf{Rule-GRPO:} The SFT model finetuned on a response set augmented with new responses, which are regenerated based on feedback from predefined rules.
  \item \textbf{\method:} The SFT model finetuned on a response set augmented with new, refined responses, which are regenerated based on structured feedback from our teacher model.
\end{itemize}

\begin{table}[t]
\centering
\small %
\setlength{\tabcolsep}{5.0pt}
\caption{Performance comparison of \method~against conventional GRPO and other feedback strategies.}
\begin{tabularx}{\linewidth}{l|ccccc|c}
\toprule
Method & NC$\uparrow$ & DAC$\uparrow$ & TTC$\uparrow$ & CF$\uparrow$ & EP$\uparrow$ & PDMS$\uparrow$ \\
\midrule
SFT & 98.5 & 95.5 & 95.3 & 100 & 81.3 & 87.4 \\
GRPO & 98.5 & 96.7 & 95.4 & 100 & 83.2 & 89.0 \\
GT-GRPO & 98.1 & 97.1 & 93.5 & 100 & 85.2 & 89.2 \\
Rule-GRPO & 98.3 & 97.3 & 94.8 & 100 & 84.5 & 89.6 \\
\rowcolor{gray!30} \textbf{\method}  & \textbf{98.9} & \textbf{98.1} & \textbf{96.0} & \textbf{100} & \textbf{85.3} & \textbf{91.0} \\
\bottomrule
\end{tabularx}
\label{table:table2}
\end{table}

\begin{figure}[t]
    \centering
    \includegraphics[width=1.0\linewidth]{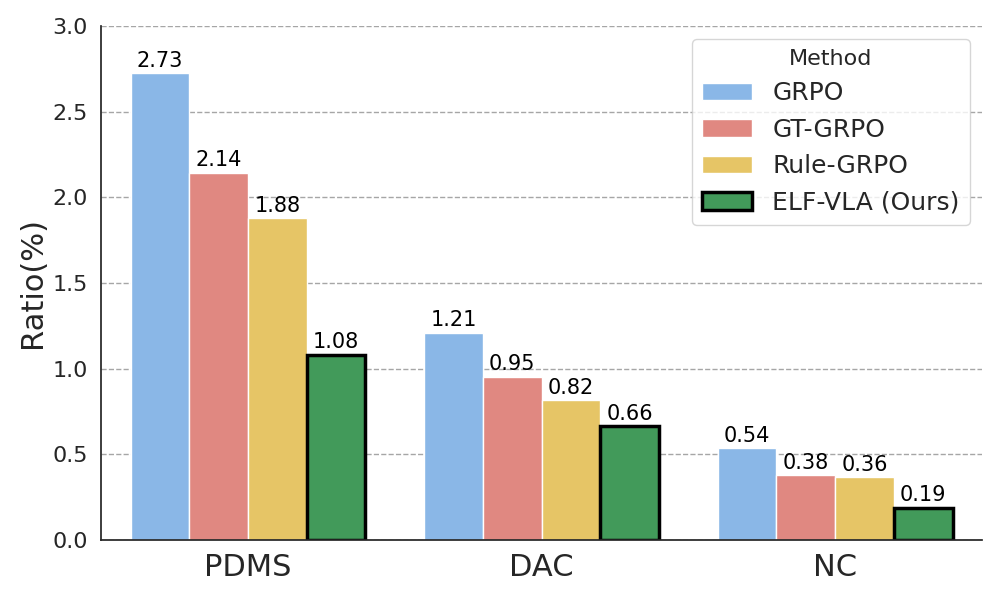}
    \caption{\textbf{Ratio of total-failure samples} measured during the RL training phase for GRPO, GT-GRPO, Rule-GRPO, and \method. A total failure indicates all rollouts for a sample failed on a specific metric (PDMS below $s$, NC of 0 and DAC of 0, respectively).}
    \label{fig:ratio_rollout}
\end{figure}

Notably, \method~achieves the best overall performance. Our method outperforms the conventional GRPO method by 2.0 PDMS. This demonstrates that by introducing structured feedback and regenerating superior, in-distribution trajectories, our approach helps the model resolve persistent failure issues. Furthermore, \method~surpasses GT-GRPO and Rule-GRPO by 1.8 and 1.4 PDMS, respectively. This highlights the distinct limitations of these two baselines. For GT-GRPO, the GT trajectories exhibit a significant distributional shift from the original VLA-generated responses. The low likelihood of these GT responses makes optimization difficult. For Rule-GRPO, the feedback from predefined rules has a limited impact on the model. This process is akin to simple self-refinement and lacks granular guidance, causing the model to fail to learn effective trajectory correction from such simplistic feedback.
In contrast, \method~utilizes the teacher model's extensive general knowledge to perform a deep, structured analysis of the original response. 
The VLA model receives this comprehensive feedback, which enables it to learn from failures and refine the trajectory. This process results in a superior, more easily optimizable refined trajectory.

\noindent \textbf{Total-Failure Ratio Analysis.} 
We analyze the failure rates during the RL training phase across these models, as shown in Fig.~\ref{fig:ratio_rollout}. Specifically, we measure the proportion of samples where all rolled-out trajectories fail simultaneously on key metrics: PDMS, DAC, and NC. As illustrated in the figure, while intermediate strategies like GT-GRPO and Rule-GRPO help reduce failure ratios, \method~demonstrates the most significant improvements across all metrics. \method~reduces the total-failure PDMS rate from 2.73\% (for GRPO) to just 1.08\%, with similarly strong reductions observed for NC and DAC. This result further validates that our method enables the model to learn from its mistakes, address the persistent failure problem, and ultimately enhance overall driving safety and robustness.

\noindent \textbf{High-Level Planning Evaluation.}
As shown in Tab.~\ref{table:table3}, our results highlight the clear advantage of \method~in high-level planning. \method~achieves the best results in both longitudinal Speed Accuracy and lateral Path Accuracy, achieving the highest overall Planning Accuracy of 80.3\%, 1.0\% higher than conventional GRPO. Moreover, compared to open-source models, \method~outperforms the significantly larger Qwen2.5-VL-72B model by 51.6\% in accuracy. This improvement stems from the teacher model providing refined meta actions, where the VLA model learns to internalize. This demonstrates that \method~can learn from failure cases to refine its high-level planning.

\begin{table}[t]
\centering
\small
\caption{Comparison of high-level planning on NAVSIM.}
\label{tab:unified_performance}
\begin{tabular}{l |cc|c}
\toprule
Method & Speed Acc.$\uparrow$ & Path Acc.$\uparrow$ & Accuracy$\uparrow$ \\
\midrule
Qwen2.5-VL-7B & 37.8 & 61.3 & 19.1 \\
InternVL3-8B & 40.9 & 58.7 & 20.1 \\
Qwen2.5-VL-32B & 46.6 & 55.3 & 27.6 \\
Qwen2.5-VL-72B & 49.4 & 62.6 & 28.7 \\
\cline{1-4}
 SFT & 84.2 & 90.7 & 79.2 \\
 GRPO & 84.3 & 90.8 & 79.3 \\
 GT-GRPO & 83.5 & 90.5 & 78.4 \\
 Rule-GRPO & 84.5 & 91.2 & 79.5 \\
\rowcolor{gray!30} \textbf{\method} & \textbf{85.8} & \textbf{92.5} & \textbf{80.3} \\
\bottomrule
\end{tabular}
\label{table:table3}
\end{table}

\begin{figure*}
    \centering
    \includegraphics[width=0.95\linewidth]{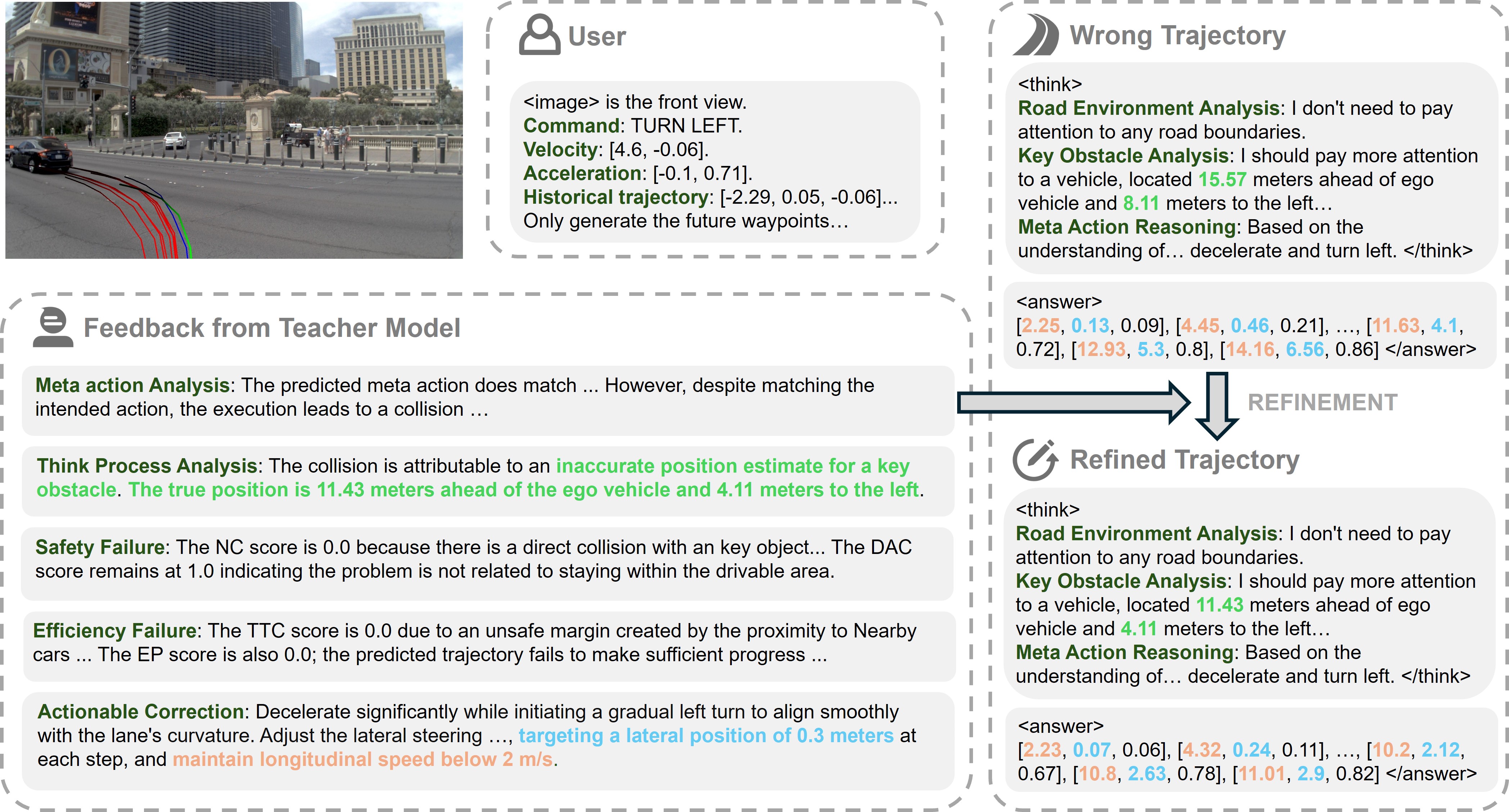}
    \caption{\textbf{Visualization of trajectory refinement process by \method~on the NAVSIM dataset.} Visualization of the initial Wrong Trajectories (\textcolor{red}{red}), the Ground Truth (\textcolor{green}{green}), and the final Refined Trajectory (\textcolor{blue}{blue}). A teacher-generated Feedback guides the refinement of a Wrong Trajectory into a Refined Trajectory. Colored text in the Feedback details the specific refinements that have been applied.}
    \label{fig:visual}
\end{figure*}

\subsection{Ablation Studies}
\noindent \textbf{On GRPO with Training Data.} We investigate the impact of training data volume and composition for RL, as shown in Tab.~\ref{table:abla_data}. Using the full 85k dataset (89.1 PDMS) or a randomly sampled 24k subset (88.9 PDMS) both yield suboptimal results. In contrast, our curated 24k dataset (24k*), guided by Sec.~\ref{grpo}, achieves the best performance of 91.0 PDMS. This suggests that the full 85k dataset is dominated by simple scenarios that provide limited learning signals, which weakens the overall gradient signal and leads to inefficient policy updates focused on already-mastered scenarios. Our curation strategy effectively distills the most valuable data. Combined with our feedback mechanism, this data allows for targeted training on these complex scenarios. This approach ultimately improves model performance and enhances training efficiency.

\begin{table}[t]
\centering
\small
\setlength{\tabcolsep}{4.2pt}
\caption{Ablation study on the number of training data in RL. $\dagger$: Randomly sampled. *: Curated as Sec.~\ref{grpo}.}
\begin{tabularx}{0.85\linewidth}{l|ccccc|c}
\toprule
Num. & NC$\uparrow$ & DAC$\uparrow$ & TTC$\uparrow$ & CF$\uparrow$ & EP$\uparrow$ & PDMS$\uparrow$ \\
\midrule
85k  & 98.5 & 96.8 & 95.3 & 100 & 83.4 & 89.1 \\
24k$\dagger$  & 98.4 & 96.8 & 95.2 & 100 & 83.1 & 88.9 \\
\rowcolor{gray!30} 24k*  & \textbf{98.9} & \textbf{98.1} & \textbf{96.0} & \textbf{100} & \textbf{85.3} & \textbf{91.0} \\

\bottomrule
\end{tabularx}
\label{table:abla_data}
\end{table}

\noindent \textbf{On GRPO with Feedback.}
Tab.~\ref{table:abla_k} analyzes our feedback components. We first vary the number of refinement responses, $k$. Optimal performance (91.0 PDMS) is achieved at $k=1$. Increasing $k$ degrades performance, dropping to 89.0 at $k=4$. This suggests that while a single targeted refinement is effective, multiple feedback-based responses may distract the policy. We also evaluate Policy Shaping (PS). Removing PS (at $k=1$) causes a significant 1.7\% drop in PDMS from 91.0 to 89.3. This confirms PS is critical for preventing training collapse and formatting errors, ensuring the model can properly learn from high-advantage, low-probability refinement trajectories.

\begin{table}[t]
\centering
\small
\vspace{-1em}
\setlength{\tabcolsep}{4pt}
\caption{Ablation study on the number of refinement responses $k$ and the use of Policy Shaping (PS).}
\begin{tabularx}{0.85\linewidth}{lc|ccccc|c}
\toprule
$k$ & PS & NC$\uparrow$ & DAC$\uparrow$ & TTC$\uparrow$ & CF$\uparrow$ & EP$\uparrow$ & PDMS$\uparrow$ \\
\midrule
4 & \ding{51} & 98.5 & 96.7 & 95.4 & 100 & 83.3 & 89.0 \\
2 & \ding{51} & 98.2 & 97.5 & 94.3 & 100 & 84.9 & 89.7 \\
1 & \ding{55} & 98.5 & 97.0 & 94.9 & 100 & 83.9 & 89.3 \\
\rowcolor{gray!30} 1 & \ding{51} & \textbf{98.9} & \textbf{98.1} & \textbf{96.0} & \textbf{100} & \textbf{85.3} & \textbf{91.0} \\

\bottomrule
\end{tabularx}
\label{table:abla_k}
\end{table}

\subsection{Visualization of Refinement Process}

Fig.~\ref{fig:visual} illustrates a qualitative example where \method~corrects a fault trajectory in a complex left-turn scenario. The initial fault trajectory (red curve) led to a potential collision, rooted in a significant misestimation of a key obstacle (predicted: 15.57m ahead, 8.11m left). Our teacher model provides structured feedback, precisely identifying this ``Think Process" error and estimating the more accurate location (11.43m ahead, 4.11m left). Concurrently, it offers actionable corrections, such as adjustments to the target lateral position and longitudinal speed. Based on this feedback, the model generates the Refined Trajectory (blue curve). The corresponding ``Key Obstacle Analysis" in the refined plan reflects this corrected perception, enabling the agent to plot a safer trajectory that successfully avoids the obstacle. More results can be found in Appendix~\ref{appendix:exp}.

\section{Conclusion}
\label{sec:conclusioni}

This paper proposes \method, a framework for explicit learning from failures.
Our approach augments the VLA policy with a powerful teacher model that produces structured diagnostic reports and identifies the underlying failure mode whenever a failure occurs. 
The policy then leverages this explicit, human-like feedback to synthesize a corrected, high-reward trajectory. 
By re-injecting these corrected samples into the RL training batch, \method~ delivers targeted gradients that enable the policy to resolve challenging scenarios where unguided exploration rarely overcomes.

The primary limitation of this method lies in its dependence on an external teacher model, which inherently bounds the student model's performance by the teacher's analytical capabilities. Furthermore, all experiments were conducted on the Navsim benchmark, a non-reactive simulation environment.
Future work will involve exploring the role of different teacher models as well as performing closed-loop evaluations on more diverse datasets.


{
    \small
    \bibliographystyle{ieeenat_fullname}
    \bibliography{main}

@String(AAAI = {AAAI})

@article{curseofrarity,
  title={Curse of rarity for autonomous vehicles},
  author={Liu, Henry X and Feng, Shuo},
  journal={nature communications},
  volume={15},
  number={1},
  pages={4808},
  year={2024},
  publisher={Nature Publishing Group UK London}
}

@article{safetycriticalsurvey,
  title={A survey on safety-critical driving scenario generation—a methodological perspective},
  author={Ding, Wenhao and Xu, Chejian and Arief, Mansur and Lin, Haohong and Li, Bo and Zhao, Ding},
  journal={IEEE Transactions on Intelligent Transportation Systems},
  volume={24},
  number={7},
  pages={6971--6988},
  year={2023},
  publisher={IEEE}
}

@article{adathinkdrive,
  title={AdaThinkDrive: Adaptive Thinking via Reinforcement Learning for Autonomous Driving},
  author={Luo, Yuechen and Li, Fang and Xu, Shaoqing and Lai, Zhiyi and Yang, Lei and Chen, Qimao and Luo, Ziang and Xie, Zixun and Jiang, Shengyin and Liu, Jiaxin and others},
  journal={arXiv preprint arXiv:2509.13769},
  year={2025}
}

@article{vlasurvey,
  title={A Survey on Vision-Language-Action Models for Autonomous Driving},
  author={Jiang, Sicong and Huang, Zilin and Qian, Kangan and Luo, Ziang and Zhu, Tianze and Zhong, Yang and Tang, Yihong and Kong, Menglin and Wang, Yunlong and Jiao, Siwen and others},
  journal={arXiv preprint arXiv:2506.24044},
  year={2025}
}

@article{e2esurvey,
  title={End-to-end autonomous driving: Challenges and frontiers},
  author={Chen, Li and Wu, Penghao and Chitta, Kashyap and Jaeger, Bernhard and Geiger, Andreas and Li, Hongyang},
  journal={IEEE Transactions on Pattern Analysis and Machine Intelligence},
  year={2024},
  publisher={IEEE}
}

@article{jiang2024senna,
  title={Senna: Bridging large vision-language models and end-to-end autonomous driving},
  author={Jiang, Bo and Chen, Shaoyu and Liao, Bencheng and Zhang, Xingyu and Yin, Wei and Zhang, Qian and Huang, Chang and Liu, Wenyu and Wang, Xinggang},
  journal={arXiv preprint arXiv:2410.22313},
  year={2024}
}

@article{jiang2025alphadrive,
  title={Alphadrive: Unleashing the power of vlms in autonomous driving via reinforcement learning and reasoning},
  author={Jiang, Bo and Chen, Shaoyu and Zhang, Qian and Liu, Wenyu and Wang, Xinggang},
  journal={arXiv preprint arXiv:2503.07608},
  year={2025}
}

@inproceedings{marcu2024lingoqa,
  title={LingoQA: Visual question answering for autonomous driving},
  author={Marcu, Ana-Maria and Chen, Long and H{\"u}nermann, Jan and Karnsund, Alice and Hanotte, Benoit and Chidananda, Prajwal and Nair, Saurabh and Badrinarayanan, Vijay and Kendall, Alex and Shotton, Jamie and others},
  booktitle={European Conference on Computer Vision},
  pages={252--269},
  year={2024},
  organization={Springer}
}

@article{tian2024drivevlm,
  title={Drivevlm: The convergence of autonomous driving and large vision-language models},
  author={Tian, Xiaoyu and Gu, Junru and Li, Bailin and Liu, Yicheng and Wang, Yang and Zhao, Zhiyong and Zhan, Kun and Jia, Peng and Lang, Xianpeng and Zhao, Hang},
  journal={arXiv preprint arXiv:2402.12289},
  year={2024}
}

@article{hwang2024emma,
  title={Emma: End-to-end multimodal model for autonomous driving},
  author={Hwang, Jyh-Jing and Xu, Runsheng and Lin, Hubert and Hung, Wei-Chih and Ji, Jingwei and Choi, Kristy and Huang, Di and He, Tong and Covington, Paul and Sapp, Benjamin and others},
  journal={arXiv preprint arXiv:2410.23262},
  year={2024}
}

@inproceedings{xing2025openemma,
  title={Openemma: Open-source multimodal model for end-to-end autonomous driving},
  author={Xing, Shuo and Qian, Chengyuan and Wang, Yuping and Hua, Hongyuan and Tian, Kexin and Zhou, Yang and Tu, Zhengzhong},
  booktitle={Proceedings of the Winter Conference on Applications of Computer Vision},
  pages={1001--1009},
  year={2025}
}

@article{qiao2025lightemma,
  title={Lightemma: Lightweight end-to-end multimodal model for autonomous driving},
  author={Qiao, Zhijie and Li, Haowei and Cao, Zhong and Liu, Henry X},
  journal={arXiv preprint arXiv:2505.00284},
  year={2025}
}

@article{zhao2025sce2drivex,
  title={Sce2drivex: A generalized mllm framework for scene-to-drive learning},
  author={Zhao, Rui and Yuan, Qirui and Li, Jinyu and Hu, Haofeng and Li, Yun and Zheng, Chengyuan and Gao, Fei},
  journal={arXiv preprint arXiv:2502.14917},
  year={2025}
}

@inproceedings{wang2025omnidrive,
  title={Omnidrive: A holistic vision-language dataset for autonomous driving with counterfactual reasoning},
  author={Wang, Shihao and Yu, Zhiding and Jiang, Xiaohui and Lan, Shiyi and Shi, Min and Chang, Nadine and Kautz, Jan and Li, Ying and Alvarez, Jose M},
  booktitle={Proceedings of the Computer Vision and Pattern Recognition Conference},
  pages={22442--22452},
  year={2025}
}

@article{fu2025orion,
  title={Orion: A holistic end-to-end autonomous driving framework by vision-language instructed action generation},
  author={Fu, Haoyu and Zhang, Diankun and Zhao, Zongchuang and Cui, Jianfeng and Liang, Dingkang and Zhang, Chong and Zhang, Dingyuan and Xie, Hongwei and Wang, Bing and Bai, Xiang},
  journal={arXiv preprint arXiv:2503.19755},
  year={2025}
}

@article{liu2025reasonplan,
  title={ReasonPlan: Unified Scene Prediction and Decision Reasoning for Closed-loop Autonomous Driving},
  author={Liu, Xueyi and Zhong, Zuodong and Guo, Yuxin and Liu, Yun-Fu and Su, Zhiguo and Zhang, Qichao and Wang, Junli and Gao, Yinfeng and Zheng, Yupeng and Lin, Qiao and others},
  journal={arXiv preprint arXiv:2505.20024},
  year={2025}
}

@inproceedings{hu2023planning,
  title={Planning-oriented autonomous driving},
  author={Hu, Yihan and Yang, Jiazhi and Chen, Li and Li, Keyu and Sima, Chonghao and Zhu, Xizhou and Chai, Siqi and Du, Senyao and Lin, Tianwei and Wang, Wenhai and others},
  booktitle={Proceedings of the IEEE/CVF conference on computer vision and pattern recognition},
  pages={17853--17862},
  year={2023}
}

@article{li2025recogdrive,
  title={ReCogDrive: A Reinforced Cognitive Framework for End-to-End Autonomous Driving},
  author={Li, Yongkang and Xiong, Kaixin and Guo, Xiangyu and Li, Fang and Yan, Sixu and Xu, Gangwei and Zhou, Lijun and Chen, Long and Sun, Haiyang and Wang, Bing and others},
  journal={arXiv preprint arXiv:2506.08052},
  year={2025}
}

@article{driver1,
  title={Drive-R1: Bridging Reasoning and Planning in VLMs for Autonomous Driving with Reinforcement Learning},
  author={Li, Yue and Tian, Meng and Zhu, Dechang and Zhu, Jiangtong and Lin, Zhenyu and Xiong, Zhiwei and Zhao, Xinhai},
  journal={arXiv preprint arXiv:2506.18234},
  year={2025}
}

@article{zhou2025autovla,
  title={AutoVLA: A Vision-Language-Action Model for End-to-End Autonomous Driving with Adaptive Reasoning and Reinforcement Fine-Tuning},
  author={Zhou, Zewei and Cai, Tianhui and Zhao, Seth Z and Zhang, Yun and Huang, Zhiyu and Zhou, Bolei and Ma, Jiaqi},
  journal={arXiv preprint arXiv:2506.13757},
  year={2025}
}

@article{shao2024deepseekmath,
  title={Deepseekmath: Pushing the limits of mathematical reasoning in open language models},
  author={Shao, Zhihong and Wang, Peiyi and Zhu, Qihao and Xu, Runxin and Song, Junxiao and Bi, Xiao and Zhang, Haowei and Zhang, Mingchuan and Li, YK and Wu, Yang and others},
  journal={arXiv preprint arXiv:2402.03300},
  year={2024}
}

@article{zhu2025internvl3,
  title={Internvl3: Exploring advanced training and test-time recipes for open-source multimodal models},
  author={Zhu, Jinguo and Wang, Weiyun and Chen, Zhe and Liu, Zhaoyang and Ye, Shenglong and Gu, Lixin and Tian, Hao and Duan, Yuchen and Su, Weijie and Shao, Jie and others},
  journal={arXiv preprint arXiv:2504.10479},
  year={2025}
}

@article{chitta2022transfuser,
  title={Transfuser: Imitation with transformer-based sensor fusion for autonomous driving},
  author={Chitta, Kashyap and Prakash, Aditya and Jaeger, Bernhard and Yu, Zehao and Renz, Katrin and Geiger, Andreas},
  journal={IEEE transactions on pattern analysis and machine intelligence},
  volume={45},
  number={11},
  pages={12878--12895},
  year={2022},
  publisher={IEEE}
}

@article{li2024hydra,
  title={Hydra-mdp: End-to-end multimodal planning with multi-target hydra-distillation},
  author={Li, Zhenxin and Li, Kailin and Wang, Shihao and Lan, Shiyi and Yu, Zhiding and Ji, Yishen and Li, Zhiqi and Zhu, Ziyue and Kautz, Jan and Wu, Zuxuan and others},
  journal={arXiv preprint arXiv:2406.06978},
  year={2024}
}

@inproceedings{liao2025diffusiondrive,
  title={Diffusiondrive: Truncated diffusion model for end-to-end autonomous driving},
  author={Liao, Bencheng and Chen, Shaoyu and Yin, Haoran and Jiang, Bo and Wang, Cheng and Yan, Sixu and Zhang, Xinbang and Li, Xiangyu and Zhang, Ying and Zhang, Qian and others},
  booktitle={Proceedings of the Computer Vision and Pattern Recognition Conference},
  pages={12037--12047},
  year={2025}
}

@article{li2025end,
  title={End-to-end driving with online trajectory evaluation via bev world model},
  author={Li, Yingyan and Wang, Yuqi and Liu, Yang and He, Jiawei and Fan, Lue and Zhang, Zhaoxiang},
  journal={arXiv preprint arXiv:2504.01941},
  year={2025}
}

@article{li2025hydra,
  title={Hydra-next: Robust closed-loop driving with open-loop training},
  author={Li, Zhenxin and Wang, Shihao and Lan, Shiyi and Yu, Zhiding and Wu, Zuxuan and Alvarez, Jose M},
  journal={arXiv preprint arXiv:2503.12030},
  year={2025}
}

@inproceedings{xing2025goalflow,
  title={Goalflow: Goal-driven flow matching for multimodal trajectories generation in end-to-end autonomous driving},
  author={Xing, Zebin and Zhang, Xingyu and Hu, Yang and Jiang, Bo and He, Tong and Zhang, Qian and Long, Xiaoxiao and Yin, Wei},
  booktitle={Proceedings of the Computer Vision and Pattern Recognition Conference},
  pages={1602--1611},
  year={2025}
}

@article{dauner2024navsim,
  title={Navsim: Data-driven non-reactive autonomous vehicle simulation and benchmarking},
  author={Dauner, Daniel and Hallgarten, Marcel and Li, Tianyu and Weng, Xinshuo and Huang, Zhiyu and Yang, Zetong and Li, Hongyang and Gilitschenski, Igor and Ivanovic, Boris and Pavone, Marco and others},
  journal={Advances in Neural Information Processing Systems},
  volume={37},
  pages={28706--28719},
  year={2024}
}

@inproceedings{sima2024drivelm,
  title={Drivelm: Driving with graph visual question answering},
  author={Sima, Chonghao and Renz, Katrin and Chitta, Kashyap and Chen, Li and Zhang, Hanxue and Xie, Chengen and Bei{\ss}wenger, Jens and Luo, Ping and Geiger, Andreas and Li, Hongyang},
  booktitle={European conference on computer vision},
  pages={256--274},
  year={2024},
  organization={Springer}
}

@article{chi2025impromptu,
  title={Impromptu VLA: Open Weights and Open Data for Driving Vision-Language-Action Models},
  author={Chi, Haohan and Gao, Huan-ang and Liu, Ziming and Liu, Jianing and Liu, Chenyu and Li, Jinwei and Yang, Kaisen and Yu, Yangcheng and Wang, Zeda and Li, Wenyi and others},
  journal={arXiv preprint arXiv:2505.23757},
  year={2025}
}

@inproceedings{qian2024nuscenes,
  title={Nuscenes-qa: A multi-modal visual question answering benchmark for autonomous driving scenario},
  author={Qian, Tianwen and Chen, Jingjing and Zhuo, Linhai and Jiao, Yang and Jiang, Yu-Gang},
  booktitle={Proceedings of the AAAI Conference on Artificial Intelligence},
  volume={38},
  number={5},
  pages={4542--4550},
  year={2024}
}

@inproceedings{ding2024holistic,
  title={Holistic autonomous driving understanding by bird's-eye-view injected multi-modal large models},
  author={Ding, Xinpeng and Han, Jianhua and Xu, Hang and Liang, Xiaodan and Zhang, Wei and Li, Xiaomeng},
  booktitle={Proceedings of the IEEE/CVF Conference on Computer Vision and Pattern Recognition},
  pages={13668--13677},
  year={2024}
}

@article{luffy,
  title={Learning to reason under off-policy guidance},
  author={Yan, Jianhao and Li, Yafu and Hu, Zican and Wang, Zhi and Cui, Ganqu and Qu, Xiaoye and Cheng, Yu and Zhang, Yue},
  journal={arXiv preprint arXiv:2504.14945},
  year={2025}
}

@article{bai2025qwen2,
  title={Qwen2. 5-vl technical report},
  author={Bai, Shuai and Chen, Keqin and Liu, Xuejing and Wang, Jialin and Ge, Wenbin and Song, Sibo and Dang, Kai and Wang, Peng and Wang, Shijie and Tang, Jun and others},
  journal={arXiv preprint arXiv:2502.13923},
  year={2025}
}

@article{zhang2025critique,
  title={Critique-grpo: Advancing llm reasoning with natural language and numerical feedback},
  author={Zhang, Xiaoying and Sun, Hao and Zhang, Yipeng and Feng, Kaituo and Lu, Chaochao and Yang, Chao and Meng, Helen},
  journal={arXiv preprint arXiv:2506.03106},
  year={2025}
}

@article{chen2024learning,
  title={Learning from natural language feedback},
  author={Chen, Angelica and Scheurer, J{\'e}r{\'e}my and Campos, Jon Ander and Korbak, Tomasz and Chan, Jun Shern and Bowman, Samuel R and Cho, Kyunghyun and Perez, Ethan},
  journal={Transactions on machine learning research},
  year={2024},
  publisher={OpenReview}
}

@article{ma2025learning,
  title={Learning What Reinforcement Learning Can't: Interleaved Online Fine-Tuning for Hardest Questions},
  author={Ma, Lu and Liang, Hao and Qiang, Meiyi and Tang, Lexiang and Ma, Xiaochen and Wong, Zhen Hao and Niu, Junbo and Shen, Chengyu and He, Runming and Li, Yanhao and others},
  journal={arXiv preprint arXiv:2506.07527},
  year={2025}
}

@article{yao2025drivesuprim,
  title={DriveSuprim: Towards Precise Trajectory Selection for End-to-End Planning},
  author={Yao, Wenhao and Li, Zhenxin and Lan, Shiyi and Wang, Zi and Sun, Xinglong and Alvarez, Jose M and Wu, Zuxuan},
  journal={arXiv preprint arXiv:2506.06659},
  year={2025}
}

@article{mtrdrive,
  title={MTRDrive: Memory-Tool Synergistic Reasoning for Robust Autonomous Driving in Corner Cases},
  author={Luo, Ziang and Qian, Kangan and Wang, Jiahua and Luo, Yuechen and Miao, Jinyu and Fu, Zheng and Wang, Yunlong and Jiang, Sicong and Huang, Zilin and Hu, Yifei and others},
  journal={arXiv preprint arXiv:2509.20843},
  year={2025}
}

@article{li2025drivevla,
  title={DriveVLA-W0: World Models Amplify Data Scaling Law in Autonomous Driving},
  author={Li, Yingyan and Shang, Shuyao and Liu, Weisong and Zhan, Bing and Wang, Haochen and Wang, Yuqi and Chen, Yuntao and Wang, Xiaoman and An, Yasong and Tang, Chufeng and others},
  journal={arXiv preprint arXiv:2510.12796},
  year={2025}
}

@article{cao2025pseudo,
  title={Pseudo-simulation for autonomous driving},
  author={Cao, Wei and Hallgarten, Marcel and Li, Tianyu and Dauner, Daniel and Gu, Xunjiang and Wang, Caojun and Miron, Yakov and Aiello, Marco and Li, Hongyang and Gilitschenski, Igor and others},
  journal={arXiv preprint arXiv:2506.04218},
  year={2025}
}
}

\clearpage
\setcounter{page}{1}

\maketitlesupplementary

In this supplementary material, we present comprehensive implementation details regarding data construction (Section~\ref{sec:data}) and reward design (Section~\ref{sec:method}), as well as additional experimental results, including ablation studies on the training pipeline and visualizations of the trajectory refinement process (Section~\ref{sec:exp}).

\section{Data Construction Details} \label{sec:data}
\subsection{Details of Pre-training Data} \label{appendix:pretrain_data}
We assembled a diverse collection of open-source driving QA datasets followed by ReCogDrive\cite{li2025recogdrive}, including DriveLM~\cite{sima2024drivelm}, LingoQA~\cite{marcu2024lingoqa}, ImpromptuVLA~\cite{chi2025impromptu}, NuScenes-QA~\cite{qian2024nuscenes}, NuInstruct~\cite{ding2024holistic}, OminiDrive~\cite{wang2025omnidrive}.








\subsection{Details of SFT Dataset}
\noindent{\textbf{CoT Construction.}} \label{cot}
Following the data construction paradigm outlined in \cite{adathinkdrive}, we generate high-quality CoT supervision by systematically synthesizing future trajectory data with scene-level semantics. In terms of dynamic entities, we filter and identify agents that actively interact with the ego vehicle based on spatio-temporal relationships. These agents are classified into three distinct categories: CIPO-1 refers to the leading vehicle in the current lane which primarily imposes longitudinal constraints; CIPO-2 includes vehicles from adjacent lanes that are inferred to merge or cut in based on lane geometry and relative velocity; and motion interaction encompasses entities whose predicted trajectories spatially intersect with the ego vehicle, indicating a high risk of collision. Regarding static elements, we leverage the NAVSIM map to reconstruct lane topology and extract critical boundary features, such as road curvature and lane centerlines, to ensure the drivable area is strictly defined. Furthermore, Qwen3-VL-32B is employed to provide fine-grained descriptions of environmental attributes, translating visual cues. By integrating these dynamic interactions, static constraints, and semantic descriptions, we curate scenarios to construct step-wise reasoning annotations that logically bridge perception, prediction, and planning. Details of CoT are shown in Fig.~\ref{fig:visual1} and Fig.~\ref{fig:visual2}.

\begin{figure*}
    \centering
    \includegraphics[width=0.95\linewidth]{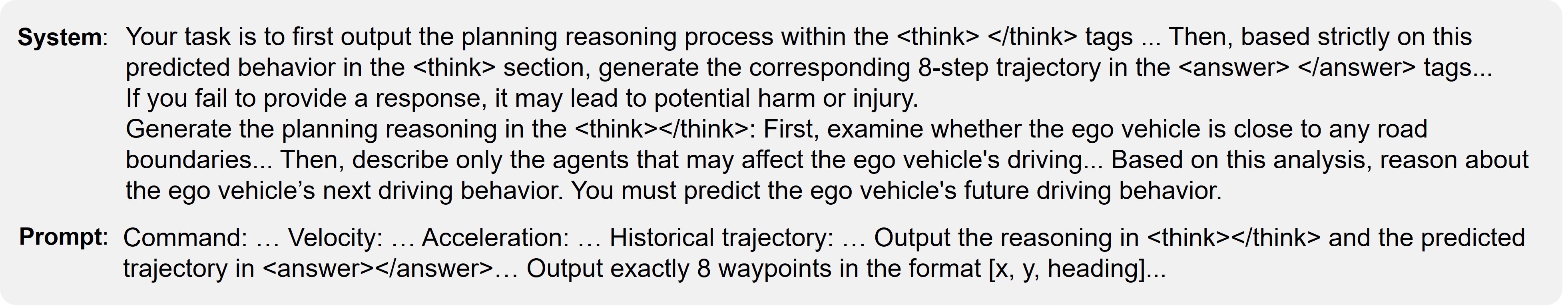}
    \vspace{-2.5mm}
    \caption{Prompt design of VLA Model (Base Inputs).}
    \vspace{-0.5em}
    \label{fig:base_prompt}
\end{figure*}

\begin{figure*}
    \centering
    \includegraphics[width=0.95\linewidth]{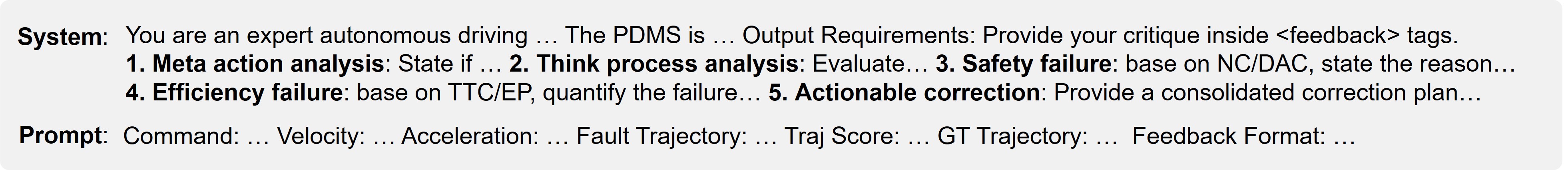}
    \vspace{-2.5mm}
    \caption{Prompt design of Teacher Model~(Feedback Inputs).}
    \vspace{-1.5em}
    \label{fig:prompt}
\end{figure*}

\noindent{\textbf{Base Inputs and Feedback Inputs Construction.}} \label{appendix:prompt}
For the base inputs, we provide the vehicle's current velocity, acceleration, and historical trajectory as prompts to help the model better predict its path, as shown in Fig.~\ref{fig:base_prompt}. For the feedback inputs, we use Qwen3VL-32B to generate structured feedback by prompting it with the Wrong Trajectory ($o_w$), Ground Truth ($o_{gt}$), detailed Navsim Metric Scores, and task requirements. These explicit inputs enable the teacher model to diagnose failure causes and generate detailed corrective guidance. Details are shown in Fig.~\ref{fig:prompt}.

\noindent{\textbf{SFT Data Construction.}}
To equip the model with a preliminary capability for trajectory refinement, we construct a feedback dataset by randomly sampling 4k correct and 4k incorrect responses from generated trajectories during the inference stage. These responses are paired with corresponding feedback based on their evaluation against a PDMS threshold $s$. Specifically, trajectory of responses scoring above $s$ are designated as ``correct" ($o_c$) and are accompanied by rule-based feedback ($f^{rule}$) to reinforce successful behaviors. Conversely, responses trajectory scoring below $s$ are labeled as ``wrong" ($o_w$) and are associated with teacher-guided feedback ($f^{teacher}$), which provides corrective feedback via inference from Qwen3-VL-32B.

\section{Method Details} \label{sec:method}
\subsection{Detailed Rewards Design in RL} \label{appendix:reward}
\noindent{\textbf{PDMS Reward.}} We input the trajectories predicted by the model into the Navsim simulator for evaluation. The simulator evaluates the driving quality of each trajectory based on several key metrics, including No At-Fault Collisions, Drivable Area Compliance, Ego Progress, Time to Collision, and Driving Comfort. It then produces a composite score, known as the \textit{Predictive Driver Model Score (PDMS)}, which serves as the reward signal for this component. This evaluation metric, \textit{PDMS}, is used for the trajectory reward $r_{\text{traj}}$, a continuous value ranging from 0 to 1.

\noindent{\textbf{Format Reward.}} The reward term $r_{fmt}$ is designed to enforce structural validity, offering a total of 1.0 point distributed evenly between two requirements. The first half (0.5 point) is awarded if the output correctly incorporates two distinct sections: $<$think$>$...$<$/think$>$ and $<$answer$>$...$<$/answer$>$. The remaining 0.5 point validates the syntax of the predicted trajectory points, ensuring the output is well-formed and machine-parsable.

\noindent{\textbf{Goal Reward.}} To promote precise alignment between the predicted endpoint and the ground truth, we employ a piecewise reward function, $r_{goal}$, calculated via the L1 distance. The specific formulation is defined as follows:

\begin{equation}
r_{goal} = 
\begin{cases} 
1 & \text{if } 0 < dis < 2  \\
0.8 & \text{if }  2  \leq dis < 4 \\
0.6 & \text{if }  4  \leq dis < 6 \\
0.4 & \text{if }  6  \leq dis < 10 \\
0.2 & \text{if }  10  \leq dis < 15 \\
0 & \text{if }  dis > 15 \\
\end{cases}
\end{equation}

\section{Experiment Details} \label{sec:exp}
\subsection{Implementation Details}  \label{appendix:training}
\noindent{\textbf{Model Architecture.}}
We use InternVL3-8B~\cite{zhu2025internvl3}, a vision-language foundation model that combines a 300M-parameter InternViT visual encoder with a 7B-parameter Qwen2.5 language model. It features a resolution-adaptive visual input mechanism that processes images by dynamically adjusting the scale and granularity of feature extraction based on the content. This design specifically applies fine-grained processing to complex regions and coarse feature extraction to simpler areas. This enables the model to maintain high visual fidelity while optimizing computational efficiency.

\noindent{\textbf{Training Parameters and Hardware Configuration.}}
The training process comprises three stages: The first stage conducts supervised fine-tuning on a large-scale, high-quality driving knowledge dataset with diverse instruction-following examples and scene-aware annotations, using 2 epochs, a batch size of 1, a learning rate of \(1 \times 10^{-5}\), 4 gradient accumulation steps, 0.05 weight decay, and 0.05 weight ratio. The second stage further fine-tunes the model on a NAVSIM planning dataset (with CoT annotations) and the feedback dataset, employing 2 epochs, a batch size of 2, a learning rate of \(4 \times 10^{-5}\), 2 gradient accumulation steps, and retaining the same weight decay and ratio (0.05 each). The third stage applies reinforcement learning with GRPO on a curated Navsim planning dataset, using a learning rate of \(2 \times 10^{-6}\), a batch size of 3, 16 gradient accumulation steps, 0.05 weight decay, 0.05 weight ratio, 8 generations, a temperature of 1.2, and 2 iterations. This stage utilizes 32 NVIDIA H20 GPUs. Additionally, we configure the threshold $s=0.8$, the policy shaping parameter $\gamma=0.1$, and set the refinement response to $k=1$. All experiments are conducted on NVIDIA H20 GPUs, using PyTorch 2.5.0 with CUDA 12.3 under Ubuntu. The first two stages are trained with 16 GPUs for approximately 2 days and 8 hours, respectively, the third stage with 32 GPUs for 18 hours. During inference, to generate CoT and trajectory, our model achieves a latency of 0.1s (accelerated by vLLM). Detailed hyperparameters are shown in Tab~\ref{tab:hyperparameters}.

\begin{figure*}
    \centering
    \includegraphics[width=0.95\linewidth]{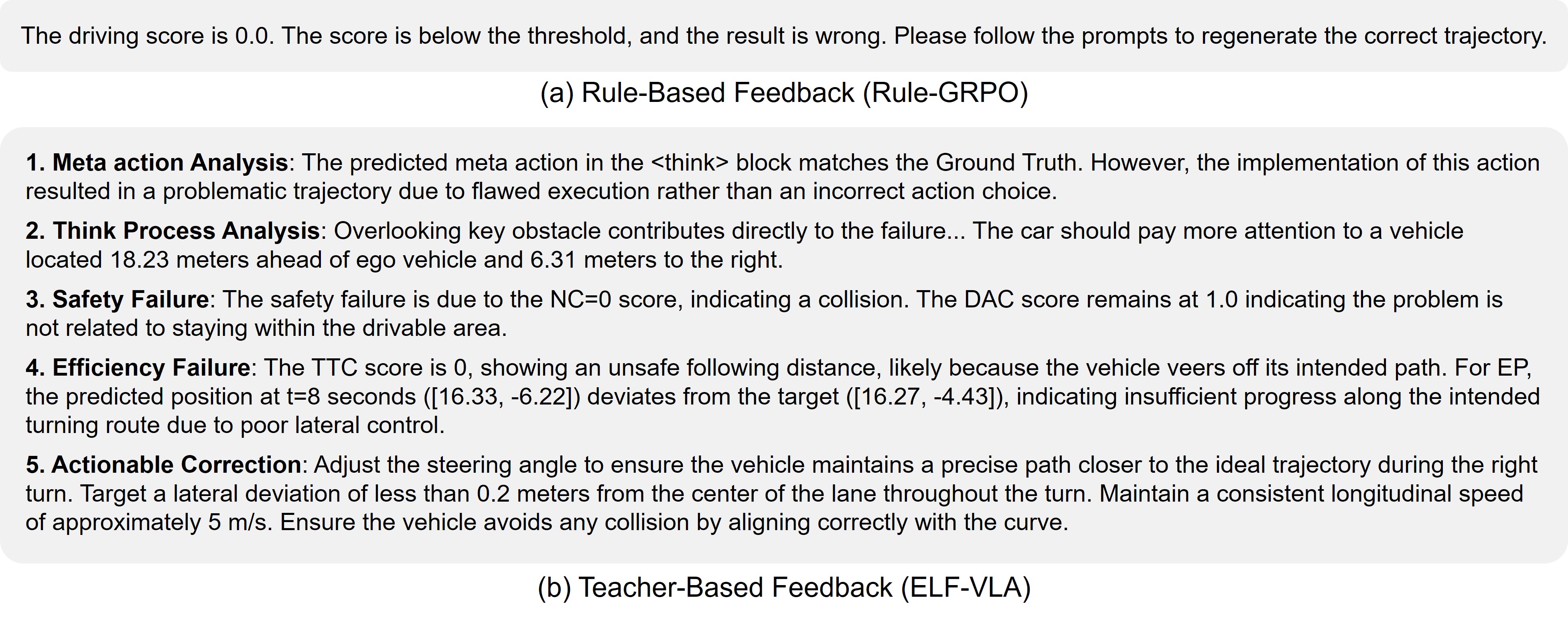}
    \vspace{-2.5mm}
    \caption{\textbf{Comparison of feedback mechanisms.} Unlike the binary signals in Rule-GRPO (a), our teacher-based feedback in \method~ (b) provides structured diagnostics and concrete, actionable strategies to guide trajectory refinement.}
    \label{fig:feedback_comparison}
    \vspace{-0.8em}
\end{figure*}

\begin{table}[h!]
\centering
\setlength{\tabcolsep}{10.0pt}
\vspace{-0.3em}
\caption{ Training hyperparameters of \method}
\vspace{-2.0mm}
\begin{tabular}{cll}
\hline
\textbf{Stage} & \textbf{Hyper-parameter} & \textbf{Value} \\ \hline
\multirow{6}{*}{Pretrain} & Epochs & 2 \\ 
 & Batch size & 1 \\ 
 & Learning rate & \(1 \times 10^{-5}\) \\  
 & Gradient accumulation steps & 4 \\ 
 & Weight decay & 0.05 \\ 
 & Weight ratio & 0.05 \\ \hline
\multirow{6}{*}{2} & Epochs & 2 \\ 
 & Batch size & 2 \\
 & Learning rate & \(4 \times 10^{-5}\) \\
 & Gradient accumulation steps & 2 \\ 
 & Weight decay & 0.05 \\
 & Weight ratio & 0.05 \\ \hline
\multirow{8}{*}{3} & Epochs & 3 \\ 
 & Batch size & 2 \\
 & Learning rate & \(2 \times 10^{-6}\) \\ 
 & Gradient accumulation steps & 16 \\ 
 & Weight decay & 0.05 \\ 
 & Weight ratio & 0.05 \\  
 & Number of generations & 8 \\ 
 & Number of iterations & 2 \\ 
 & Temperature & 1.2 \\  
 & Threshold($s$) & 0.8 \\ 
 & Number of refiniements & 1 \\ 
 & Policy shaping weight($\gamma$) & 0.1 \\ 
 \hline
\end{tabular}
\vspace{-1.5em}
\label{tab:hyperparameters}
\end{table}

\noindent{\textbf{Metric.}} \label{navsim_metric}
To evaluate trajectory prediction within the NAVSIM benchmark, we adopt the Predictive Driver Model Score (PDMS) for \textbf{NAVSIMv1}~\cite{dauner2024navsim} and the Extended Predictive Driver Model Score (EPDMS) for \textbf{NAVSIMv2}~\cite{cao2025pseudo} as our primary closed-loop planning metrics.

For NAVSIMv1, PDMS integrates five sub-metrics: No At-Fault Collision (NC), Drivable Area Compliance (DAC), Time-to-Collision (TTC), Comfort (C), and Ego Progress (EP) to produce a comprehensive closed-loop planning score. Its calculation formula is defined as follows:
\begin{equation}
PDMS = NC \times DAC \times \left( \frac{5\times EP + 5\times TTC + 2\times C}{12} \right),
\end{equation}

For NAVSIMv2,  EPDMS metric includes several components categorized as penalties or weighted subscores. Its key metrics are No at-Fault Collision (NC), Drivable Area Compliance (DAC), Driving Direction Compliance (DDC), Traffic Light Compliance (TLC), Ego Progress (EP), Time to Collision (TTC), Lane Keeping (LK), History Comfort (HC), and Extended Comfort (EC). Its calculation formula is defined as follows:
\begin{equation}
\begin{split}
    EPDMS ={} & NC \times DAC \times DDC \times TLC \times \\
              & \left( \frac{5 EP + 2 LK + 2 HC + 5 TTC + 2 EC}{16} \right).
\end{split}
\end{equation}

\noindent\textbf{Comparison of Feedback Mechanisms.}  \label{appedix:fb_mechanisms}
As outlined in Sec~\ref{exp_compa}, the distinction between Rule-GRPO and \method~ lies fundamentally in the mechanism and granularity of the feedback. 
Rule-GRPO relies on static heuristics that provide binary feedback which simply indicates correctness or failure, subsequently referencing the ground truth to guide the regeneration of the correct answer. 
In contrast, \method~ generates online, instance-specific feedback. By leveraging the fine-grained metrics within PDMS, it analyzes the specific causes of error for each faulty trajectory, providing detailed reasoning and constructive guidance to steer the correction process. 
As illustrated in Fig.~\ref{fig:feedback_comparison}, rule-based feedback offers limited constraints, and \method's feedback delivers comprehensive diagnostics for precise trajectory optimization.

\begin{figure*}
    \centering
    \includegraphics[width=0.95\linewidth]{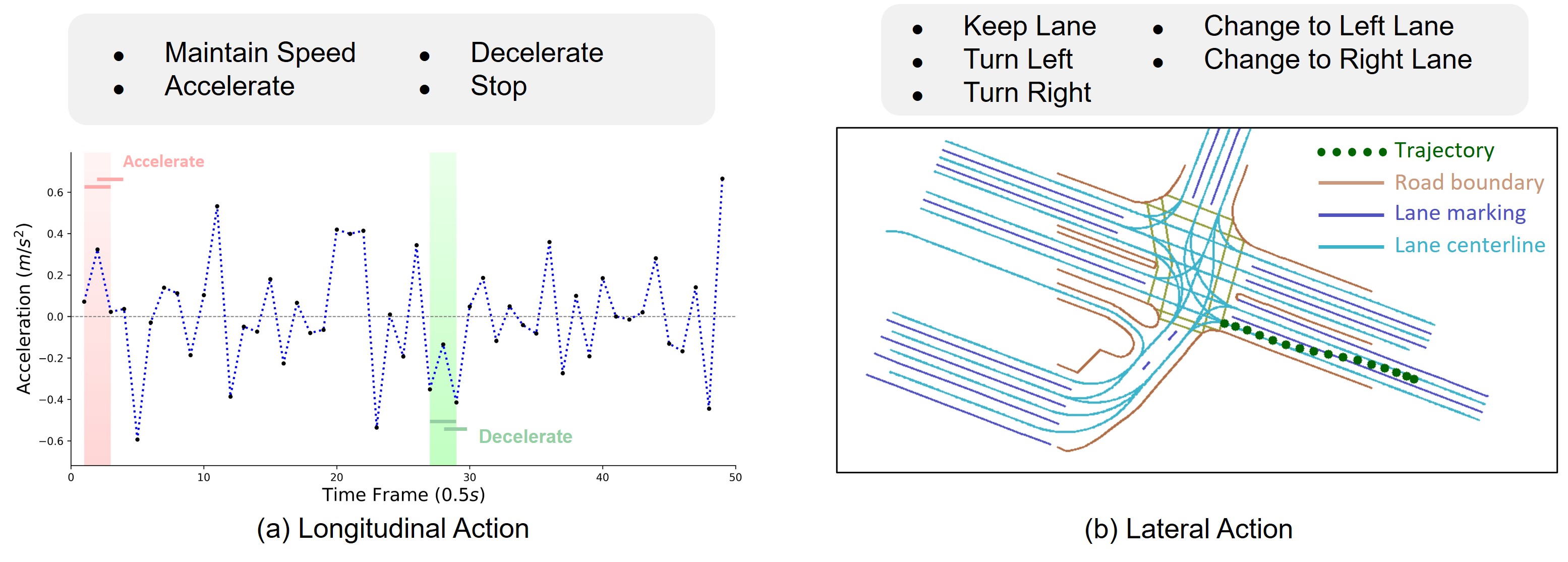}
    \vspace{-1em}
    \caption{\textbf{Illustration of high-level actions and ground truth generation.} The top panels list the discrete categories for longitudinal and lateral planning. The bottom panels demonstrate the labeling criteria: longitudinal states are determined by sliding-window acceleration analysis, while lateral behaviors are identified based on the relationship between the vehicle's trajectory and map topology.}
    \label{fig:meta}
    \vspace{-1em}
\end{figure*}

\noindent\textbf{High-Level Planning Accuracy.}  \label{appendix:high_level}
To evaluate the model's decision-making capabilities, we employ the high-level planning accuracy metric, which assesses the precision across two distinct dimensions: longitudinal speed and lateral path. The longitudinal dimension classifies vehicle behavior into four discrete states: accelerate, decelerate, maintain speed, and stop. The lateral dimension encompasses five directional maneuvers: keep lane, turn left, turn right, change to the left lane, and change to the right lane. As illustrated in Fig.~\ref{fig:meta}, the ground truth labels are derived directly from the trajectories. Specifically, longitudinal states are determined by calculating acceleration via a sliding window. For lateral maneuvers, we construct a road topology graph to identify the specific action based on the alignment between the future path and the lane structure.

\subsection{More Ablation Studies} \label{appendix:exp}
\noindent \textbf{On Training Pipeline.}
Tab.~\ref{table:ablation_train} presents the ablation results of the three-stage training pipeline for \method. Using only NAVSIM trajectory data for SFT, the model achieves a PDMS of 85.3. Adding pretraining on a large-scale driving QA dataset boosts the score to 87.4 PDMS, representing a 2.1 increase. Incorporating Feedback-Grpo further improves performance to 91.0 PDMS, a gain of 3.6. These results demonstrate that both pretraining and the Feedback-Grpo strategy play a crucial role in enhancing the model's understanding and reasoning capabilities.

\vspace{1em}

\begin{table}[t]
\centering
\small %
\setlength{\tabcolsep}{4.5pt}
\caption{\textbf{Ablation Study on \method~Components.} We evaluate the effect of pre-training, supervised fine-tuning, and reinforcement learning on driving performance using NAVSIM benchmark.}
\begin{tabularx}{\linewidth}{l|ccccc|c}
\toprule
Model & NC$\uparrow$ & DAC$\uparrow$ & TTC$\uparrow$ & CF$\uparrow$ & EP$\uparrow$ & PDMS$\uparrow$ \\
\midrule
SFT & 98.5 & 93.4 & 95.1 & 100 & 78.8 & 85.3 \\
Pre+SFT & 98.5 & 95.5 & 95.3 & 100 & 81.2 & 87.4 \\
\rowcolor{gray!30} Pre+SFT+RL & \textbf{98.9} & \textbf{98.1} & \textbf{96.0} & \textbf{100} & \textbf{85.3} & \textbf{91.0} \\
\bottomrule
\end{tabularx}
\label{table:ablation_train}
\end{table}

\noindent \textbf{On Pre-training for Feedback-GRPO.}
We conducted an ablation study excluding pre-training datasets (see Tab.~\ref{table:pretrain}). Even without pre-training, ELF-VLA achieves 90.0 PDMS, surpassing the baseline (86.9 PDMS) by +3.1. This confirms that the performance gains are primarily driven by our GRPO design rather than just pretrain data.

\begin{table}[t]
\centering
\setlength{\tabcolsep}{1.2pt}
\small
\caption{\textbf{Performance of \method~without pre-training.} ``w/o'' denotes ``without''.}
\begin{tabular}{l|ccccc|c}
    \toprule
    Method & NC$\uparrow$ & DAC$\uparrow$ & TTC$\uparrow$ & CF$\uparrow$ & EP$\uparrow$ & PDMS$\uparrow$ \\
    \midrule
    InternVL3-8B(w/o pretrain) & 97.9 & 93.9 & \textbf{94.4} & \textbf{100} & 82.8 & 86.9 \\
\rowcolor{gray!30}    ELF-VLA-8B(w/o pretrain)   & \textbf{98.1} & \textbf{97.0} & \textbf{94.4} & \textbf{100} &  \textbf{86.2} & \textbf{90.0} \\
    \bottomrule
\end{tabular}
\label{table:pretrain}
\end{table}

\noindent \textbf{On Feedback Threshold $s$.}
We investigate the sensitivity of the threshold $s$, which serves as the criterion for triggering the teacher model's feedback mechanism (as formulated in Sec.~\ref{formulation}). Specifically, the teacher is invoked to generate refinement guidance only when the model's response score falls below $s$. 
As shown in Tab.~\ref{table:ablation2}, the model achieves peak performance at $s=0.8$. 
Lower thresholds (e.g., 0 and 0.5) yield suboptimal results, primarily because they restrict the scope of correction to complete failures (score 0), neglecting marginally poor samples that still require improvement. By setting $s=0.8$, we effectively expand the refinement scope to capture and correct these suboptimal cases. 
Conversely, aggressively increasing the threshold to 0.9 leads to a performance degradation. This suggests that samples scoring in the range of $[0.8, 0.9)$ are already sufficiently high-quality. Forcing refinement on these valid responses yields no positive gain and may instead introduce noise, causing the optimization process to diverge from the optimal policy. 
Consequently, our \method~ employs $s=0.8$ to balance correction coverage and training stability.

\begin{table}[t]
\centering
\small %
\setlength{\tabcolsep}{5.5pt}
\caption{Ablation Study on the Feedback Threshold $s$.}
\begin{tabularx}{\linewidth}{l|ccccc|c}
\toprule
Threshold & NC$\uparrow$ & DAC$\uparrow$ & TTC$\uparrow$ & CF$\uparrow$ & EP$\uparrow$ & PDMS$\uparrow$ \\
\midrule
0 & 98.4 & 97.1 & 94.3 & 100 & 84.7 & 89.4 \\
0.5 & 98.6 & 97.4 & 95.3 & 100 & 84.9 & 90.1 \\
0.9 & 98.4 & 97.3 & 94.9 & 100 & 84.8 & 89.8 \\
\rowcolor{gray!30} \textbf{0.8} & \textbf{98.9} & \textbf{98.1} & \textbf{96.0} & \textbf{100} & \textbf{85.3} & \textbf{91.0} \\
\bottomrule
\end{tabularx}
\label{table:ablation2}
\end{table}

\subsection{Visualization of Refinement Process}

\begin{figure*}
    \centering
    \includegraphics[width=0.95\linewidth]{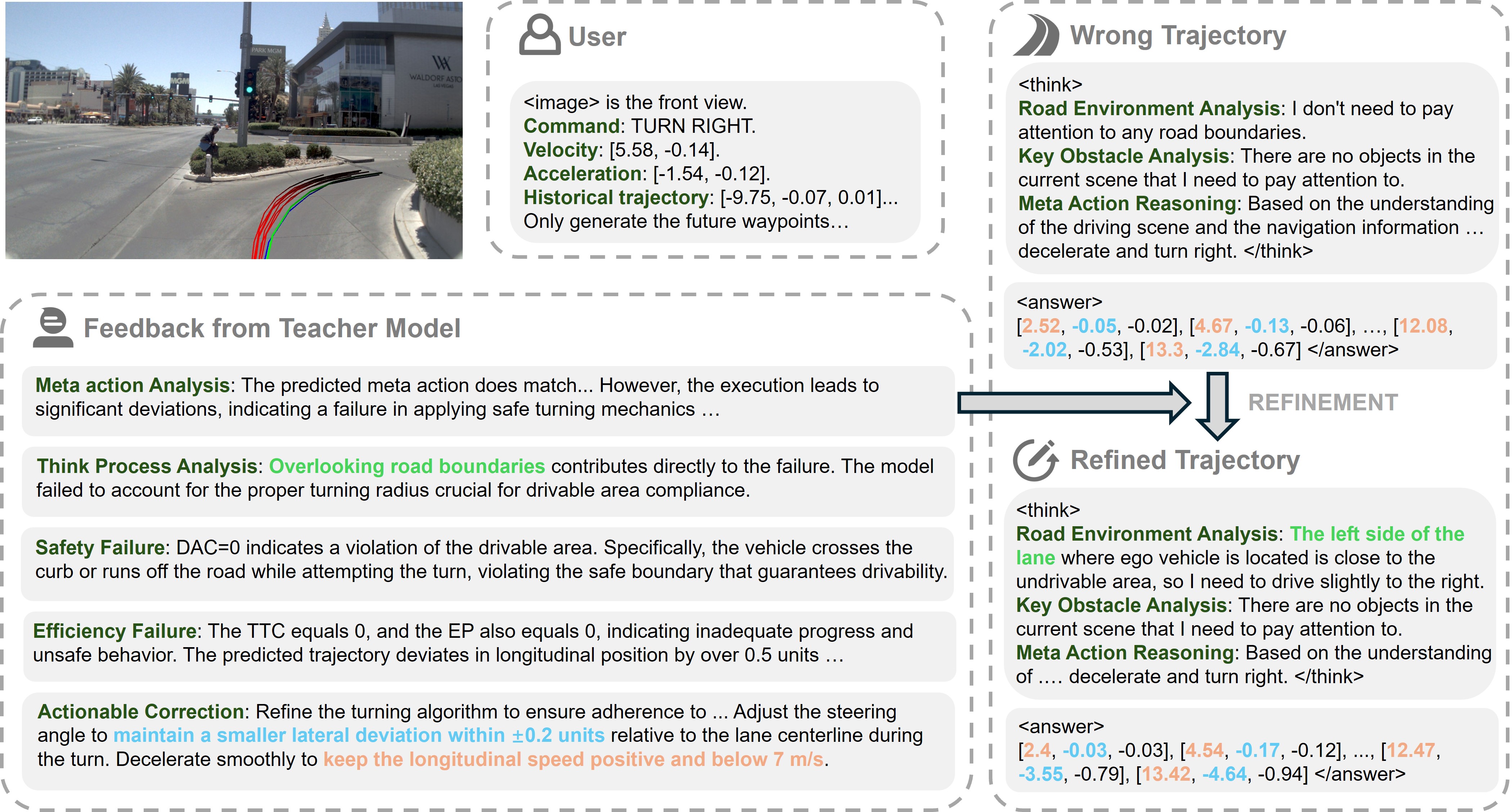}
    \caption{\textbf{Visualization of trajectory refinement process by \method~on the NAVSIM dataset.} Visualization of the initial Wrong Trajectories (\textcolor{red}{red}), the Ground Truth (\textcolor{green}{green}), and the final Refined Trajectory (\textcolor{blue}{blue}).}
    \label{fig:visual1}
\end{figure*}

\begin{figure*}
    \centering
    \includegraphics[width=0.95\linewidth]{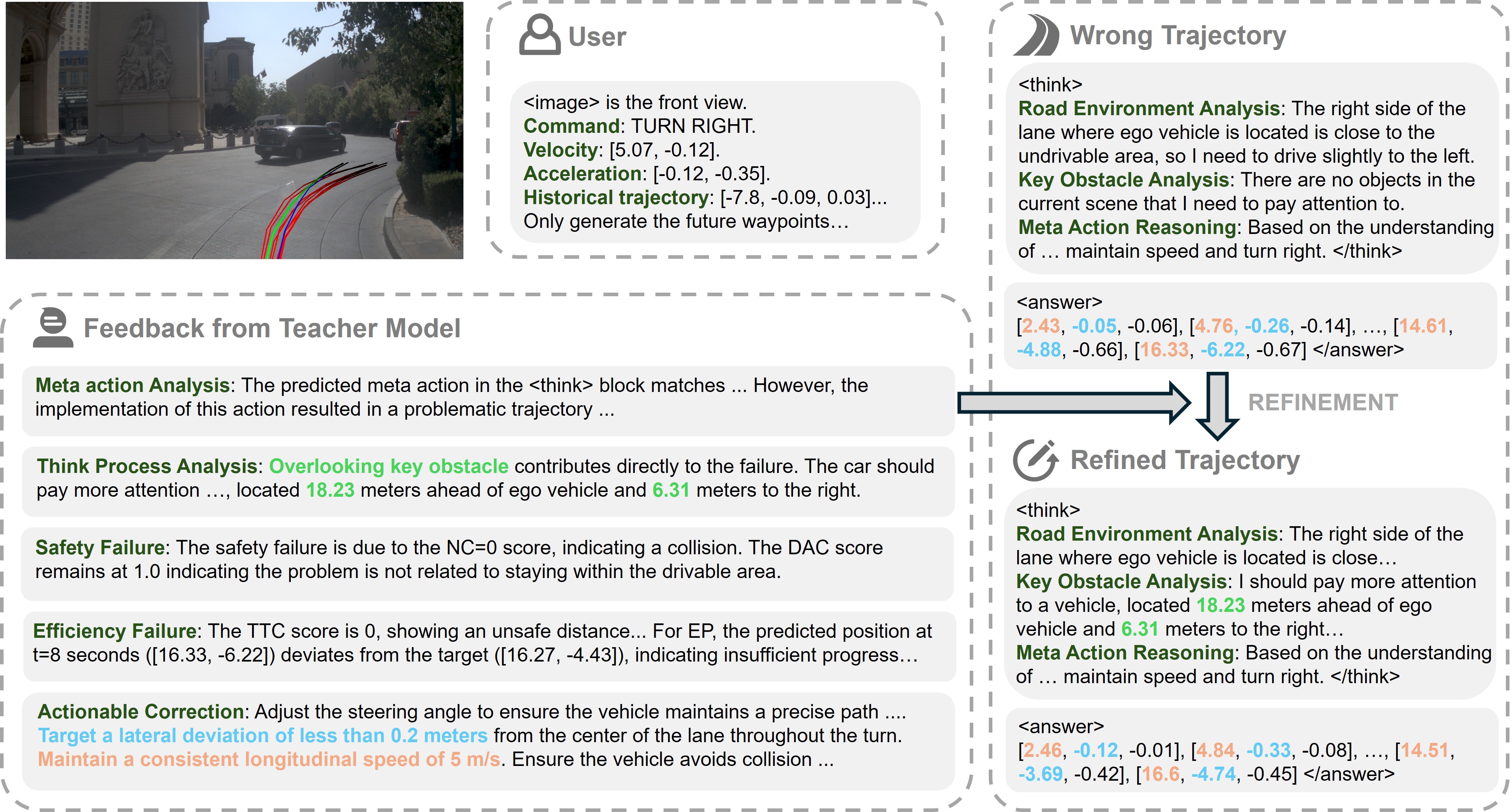}
    \caption{\textbf{Visualization of trajectory refinement process by \method~on the NAVSIM dataset.} Visualization of the initial Wrong Trajectories (\textcolor{red}{red}), the Ground Truth (\textcolor{green}{green}), and the final Refined Trajectory (\textcolor{blue}{blue}).}
    \label{fig:visual2}
\end{figure*}

Fig.~\ref{fig:visual1} and Fig.~\ref{fig:visual2} presents additional qualitative examples demonstrating the efficacy of \method~in complex scenarios. As illustrated, our structured feedback mechanism plays a critical role in the refinement loop. 
First, it validates the correctness of high-level planning decisions. 
Second, and crucially, it rectifies the intermediate CoT reasoning process. This step effectively mitigates the risk of error accumulation, ensuring that the CoT serves its intended purpose of enhancing trajectory prediction rather than introducing hallucinations or cascading faults. To achieve this, we leverage the spatial reasoning capabilities of the teacher model (Qwen3-VL-32B) to provide precise localization of key obstacles. 
Furthermore, aided by the granular scoring within PDMS, we conduct a detailed diagnosis of failure modes, specifically categorized into safety failures and efficiency failures. This comprehensive analysis culminates in the generation of concrete, actionable correction strategies to refine the final output.

\end{document}